\RequirePackage{iftex}

\long\def\pdfinfo#1{}
\documentclass[11pt,letterpaper]{article}
\usepackage[T1]{fontenc}
\usepackage{lmodern}
\usepackage[margin=1in]{geometry}
\usepackage{microtype}
\usepackage{graphicx}
\usepackage{amsmath,amssymb}
\usepackage{booktabs,array}
\usepackage{float}
\usepackage[section]{placeins}
\usepackage[round,authoryear]{natbib}
\usepackage[hyphens]{url}
\usepackage{caption}
\usepackage{listings}
\usepackage{xcolor}
\usepackage[hidelinks]{hyperref}
\hypersetup{pdftitle={Inside the Trading Loop: Markets, Models, and Moves with LLM Agents},
 pdfauthor={Renee Jia, Di Mu, Yao Yu, Tianrui Liu},pdfsubject={Multi-agent trading system, portfolio design, benchmark returns, and execution trade-offs}}
\newcolumntype{L}[1]{>{\raggedright\arraybackslash}p{#1}}
\newcommand{\Lz}{L0}\newcommand{\Lo}{L1}\newcommand{\Lt}{L2}\newcommand{\Lth}{L3}
\newcommand{\model}{Claude Haiku~4.5}
\newcommand{\pp}{\,pp}
\newcommand{\SPLITTEST}{98\%}\newcommand{\SPLITTRAIN}{98\%}

\newcommand{\authorrowbreak}{\par\vspace{15pt}}
\newcommand{\authorentry}[3]{%
  \begin{minipage}[t]{0.475\textwidth}
    \raggedright
    {\normalsize\bfseries #1\par}
    \vspace{2pt}
    {\small #2\par}
    \vspace{1pt}
    {\footnotesize\ttfamily #3\par}
  \end{minipage}%
}
\makeatletter
\renewcommand{\@maketitle}{%
  \newpage
  \begingroup
    \setlength{\parindent}{0pt}
    {\raggedright\fontsize{20}{24}\selectfont\bfseries\@title\par}
    \vspace{22pt}
    \@author\par
  \endgroup
  \vspace{18pt}
}
\makeatother
\title{Inside the Trading Loop:\\
Markets, Models, and Moves with LLM Agents}
\author{%
  \authorentry{Renee Jia\thanks{Correspondence: \texttt{reneejia@r2m.ai}.}}%
    {R2M AI, Cornell University}{reneejia@r2m.ai}%
  \hfill
  \authorentry{Di Mu}{R2M AI, Carnegie Mellon University}{dimu@r2m.ai}%
  \authorrowbreak
  \authorentry{Yao Yu}{Shanghai Jiao Tong University}{csyuyao@sjtu.edu.cn}%
  \hfill
  \authorentry{Tianrui Liu}{University of California San Diego}{til028@ucsd.edu}%
}
\date{}
\begin{document}
\maketitle

\begin{abstract}
We built a hybrid multi-agent trading system that combines large language model (LLM) analysts with momentum-based portfolio construction and rule-based execution. Company-news and macroeconomic agents return structured reports to a coordinator that sets stock weights and cash reserves. We evaluate the portfolio engine using its broker-recorded paper account, which operated with deterministic fallbacks, and compare LLM-enabled configurations through retrospective replays. From March 11 to August 28, 2026, the account returned 34.5\%, compared with 14.4\% for the S\&P~500 fund SPY over the same dates. This raw return advantage came with a larger maximum drawdown. Reconstructed policies also finish above matched SPY and QQQ references when assessments are used in the same session or deferred to the next. Design comparisons examine cash authority, selective invocation, and response contracts. A constant-cash control gives a return estimate close to stance-based sizing, and shorter report contracts improve observed parsing but can change recommendations.
\end{abstract}

\section{Introduction}
Building an automated portfolio requires a policy that connects market information to positions and orders. Stock selection, cash reserves, and rebalancing must be considered together because each trade changes the starting state of the next decision. Large language models (LLMs) can supply company and macroeconomic assessments; the portfolio policy must specify how those assessments affect capital.

Recent trading agents use specialized analytical roles, tools, and reflection to process financial information \citep{tradingagents2024,finagent2024}. News-based prediction studies examine whether language contains useful return information \citep{lopezlira2023}. In our system, analyst roles must share a portfolio budget and an execution policy. We therefore make the rules for combining their reports explicit, including which fields can change stock weights or total exposure.

We built the system around a concentrated momentum portfolio, a company-news analyst, and a macroeconomic analyst. A rule-based coordinator combines their reports with numerical factors to produce stock weights and a cash target. Momentum ranks establish the main allocation pattern, while composite scores provide a smaller adjustment. The execution layer applies versioned rebalancing rules and feeds resulting holdings into the next run. By separating these stages, we can vary analyst inputs and allocation authority while keeping the core portfolio engine fixed.

We evaluate the system with two sources of evidence. A broker-recorded paper account, operated with deterministic fallbacks, finishes the observed window above SPY, QQQ, and an initial-stock-basket reference, with a pronounced drawdown along the way. Retrospective replays compare agent-enabled configurations under the original execution abstraction and a next-session assessment rule. The account record measures portfolio performance; the replays test the analyst integrations.

We contribute an implemented trading architecture and an evaluation of its portfolio and integration policies. Alongside complete benchmark return paths, controlled comparisons test how allocation authority, request selection, and report contracts change portfolio behavior.

\section{Related Work}
\paragraph{LLM trading systems.}
TradingAgents assigns analytical and trading roles to collaborating LLM agents \citep{tradingagents2024}. FinAgent combines market information, tools, memory, and reflection within a trading agent \citep{finagent2024}. Headline-based studies evaluate language directly as a return-prediction signal \citep{lopezlira2023}. We implement two specialist analyst roles and connect their reports to a fixed portfolio coordinator. Our evaluation concerns the resulting portfolio and integration policies.

\paragraph{Portfolio construction.}
Momentum research studies the continuation of relative past returns \citep{jegadeesh1993momentum}, while portfolio selection treats allocation and risk jointly \citep{markowitz1952portfolio}. We implement a concentrated momentum policy through explicit selection, weighting, concentration, and cash rules. We evaluate these chosen rules against market references, reporting risk alongside return. Their coefficients are design settings, not estimates of an optimal allocation.

\paragraph{System evaluation and selective computation.}
Downstream dependencies, feedback, simple controls, and integration tests are established concerns in machine-learning systems \citep{sculley2015debt,breck2017mltest}. Selective prediction and model routing study coverage, quality, and computational cost \citep{geifman2019selectivenet,frugalgpt2023,routellm2024}. We connect call selection to an explicit portfolio decision and measure the allocations preserved. Work on look-ahead bias and historical trading evaluation informs the separation between account records, reconstructed policies, and prospective evidence \citep{glasserman2023,profitmirage2025,alphaillusion2026}.

Behavioral interventions provide a further reason to inspect the path from information to action. \citet{jiamu2026cot} alter an intermediate reasoning step before continuation; \citet{jiawu2026metadata} vary visible metadata while holding the task fixed. Our design comparisons instead change analyst fields or the rules that read them, and observe the resulting portfolio decisions.

\section{System Design: From Market Context to Portfolio Actions}
\label{sec:system}
\begin{figure}[htbp]
\centering
\includegraphics[width=\linewidth]{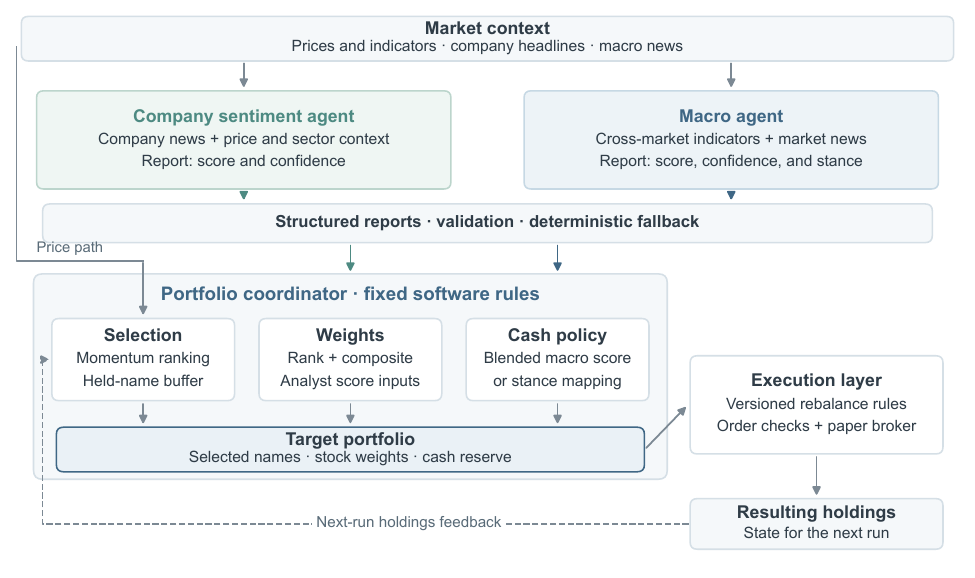}
\caption{The trading-system architecture. Market context supports two specialist analyst roles and a separate numerical selection path. Structured reports enter a deterministic coordinator, which produces a target portfolio for the execution layer. Dashed feedback carries holdings into later selection. The diagram shows the agent-enabled architecture; the recorded account used deterministic fallbacks during the study window.}
\label{fig:arch}
\end{figure}

We organized the trading loop into analyst, portfolio, and execution stages (Figure~\ref{fig:arch}). The analyst modules return structured assessments. We implemented a rule-based coordinator to combine analyst fields with numerical factors and construct a target portfolio. The execution layer converts the target into orders and passes resulting holdings to the next run. Each interface specifies what information is available and which decisions it can change.

\subsection{A Concentrated Portfolio with Controlled Adjustments}
We use a concentrated momentum portfolio as the system's investment policy. The engine ranks the eligible universe by a blend of one- and three-month momentum and targets ten stocks. This concentrates capital in the leading names while leaving the book exposed to shared stock and sector movements. We cap individual target positions to limit single-name concentration; the policy has no sector-neutrality constraint or volatility target.

We separate the selection signal from the weighting inputs. Momentum determines the selected set; within that set, we combine a momentum-rank value with a squared composite score. For stock $i$, the rule is $0.95r_i+0.05(c_i/100)^2$, where larger $r_i$ denotes stronger momentum and $c_i$ is a composite on a 0--100 scale. The composite combines numerical analyzers and sentiment, with component-specific confidence adjustments. The coordinator scales weights to the invested fraction, caps individual target positions at 25\%, and redistributes excess capital.

\begin{figure}[htbp]
\centering
\includegraphics[width=\linewidth]{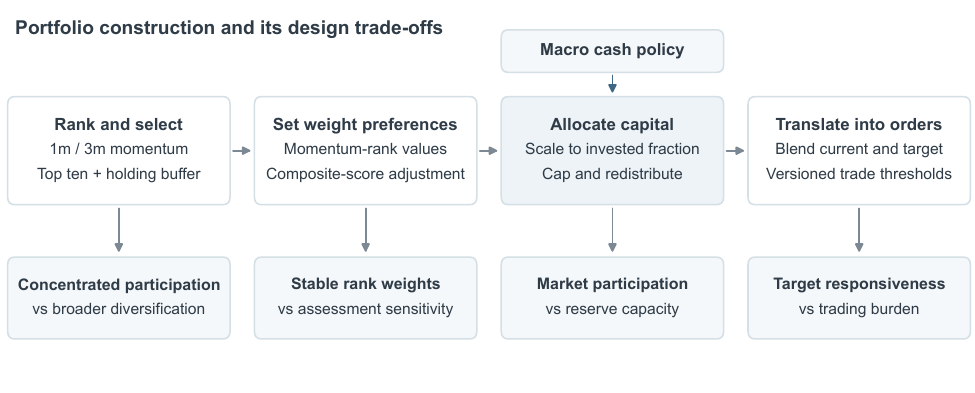}
\caption{From candidates to orders. Selection defines the opportunity set; rank and composite values establish relative weights; the cash policy sets the invested budget; and execution compares targets with current holdings. The lower row states the associated design trade-offs. These are properties and interpretations of the rules, not a decomposition of their return contributions.}
\label{fig:portfoliopolicy}
\end{figure}

We use ranks for the main weight component so that allocation depends on relative ordering without treating past price changes as calibrated expected returns. This choice gives up sensitivity to the magnitude of differences between names. The smaller composite adjustment admits other evidence while limiting how far it can move the allocation. Both coefficients act before normalization. Figure~\ref{fig:portfoliopolicy} connects these choices to the construction sequence.

In a later strategy version, we added a holding buffer: an existing name can retain a slot while its momentum rank remains within the top fifteen. The different entry and retention conditions are intended to limit replacements caused by small rank changes. They can also delay a response to a deteriorating ranking. Because the selector now reads holdings, earlier allocation and execution decisions can affect later membership.

\subsection{Specialist Reports and Explicit Allocation Authority}
We implemented the two analyst roles using \model{}, with separate task contexts, response fields, and deterministic fallbacks. The company sentiment agent reads headlines with price and sector context, returning a score and confidence. The macro agent reads cross-market indicators and market news, returning a score and an action label, or \emph{stance}. Using one base model keeps the service choice fixed; the task and context distinguish the roles.

We assign each report a defined role in the coordinator. In the advisory configuration, we use sentiment in the composite for stock weights. Its nominal coefficient within that composite is 0.15. The macro score is blended equally with a numerical macro score and also enters the trend analyzer. The blended score affects both weights and cash. The stance remains a reported recommendation. This arrangement gives the agents defined input channels while preserving a common selection and allocation procedure.

\begin{figure}[htbp]
\centering
\includegraphics[width=\linewidth]{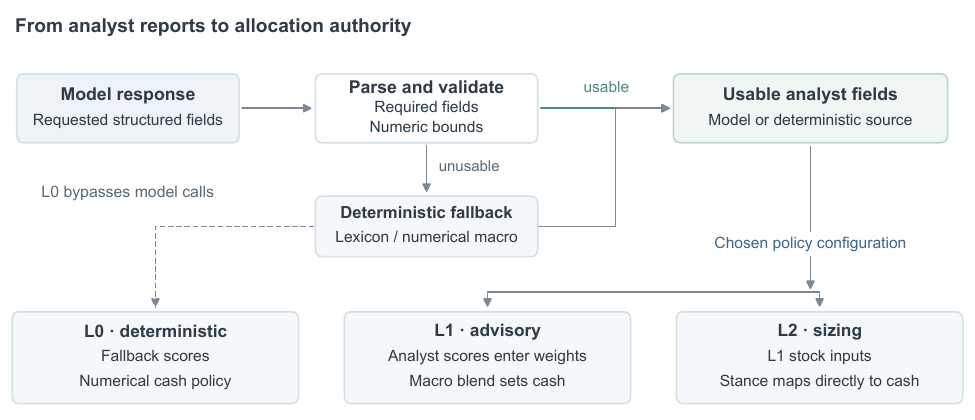}
\caption{Report handling and allocation authority. Usable model fields or deterministic fallbacks reach a specified portfolio policy. \Lz{} bypasses model calls. \Lo{} uses analyst scores, while \Lt{} additionally maps the macro stance to cash. Parsing and numeric handling are component-specific (Appendix~\ref{app:coercion}). Fallback preserves an allocation path; it does not establish equivalent information quality.}
\label{fig:agentprotocol}
\end{figure}

We implemented three allocator configurations to vary analyst authority (Figure~\ref{fig:agentprotocol}). The deterministic baseline, \Lz{}, uses fallbacks for both roles. The advisory arm, \Lo{}, enables analyst scores at the interfaces just described. The sizing arm, \Lt{}, retains the advisory stock inputs but replaces score-based cash with a fixed stance-to-cash mapping. This last change delegates an exposure decision, so it creates a different portfolio policy even when the reports themselves are unchanged.

We keep coordination in software to make the allocation policy inspectable and to control how analyst reports affect capital. The cost is that information outside the accepted fields cannot alter the portfolio. An explanation cannot override the selector, and a stance affects cash only in the sizing configuration. The implemented system has neither peer dialogue nor a learned coordinator.

\subsection{Cash Reserves and Stateful Execution}
We use a discrete reserve policy to connect the macro assessment to market exposure. The advisory allocator rounds the blended score to one decimal and maps it to four cash levels, with a maximum reserve of 20\%. The target therefore remains mostly invested. Discrete levels make the policy explicit but concentrate its sensitivity near boundaries. For the sizing configuration, we replace this score rule with a stance-to-cash map that permits a wider action set.

We implemented execution as a state-dependent transition from current holdings to the target. The executor blends that adjustment, applies minimum-trade conditions, and processes sells before buys. Small changes can remain untraded, while partial exits can leave residual holdings. The resulting book can therefore contain more names than the ten-stock target. This distinction is important when interpreting concentration, turnover, and the feedback into future selection.

Execution also changed during the historical window. Earlier versions used partial exits and permitted margin. A later version combined the rank buffer with full exits and a larger minimum-trade condition; the final version added a cash-budget guard. Faster adjustment would track the target more closely but could increase trading burden. The observed version sequence is reconstructed rather than treated as one uniform execution policy. Appendix~\ref{app:faith} reports the resulting differences between the account and the replay.

\subsection{Response Contracts and Selective Calls}
We require machine-readable fields at the analyst interface and implement a deterministic fallback for unusable replies. The parsers handle scores and confidence according to component-specific numeric rules; they do not execute free-text sizing guidance. Explanatory fields make reports easier to inspect, but compete for the response budget needed by the allocation fields. We test shorter contracts for both parsing reliability and changes in recommendations.

We also evaluate an optional call gate using information available before a macro request. Distance from the numerical macro score to a cash threshold indicates when a model-induced shift could change the reserve. We compare this rule with input-based date selection. The direct readout is the share of the full advisory arm's allocation distance occurring on selected dates. Separate replays gate both roles or macro alone. These are offline comparisons; their trading consequences have not been measured prospectively.

\section{Evaluation Protocol}
\label{sec:method}
\subsection{Account Records and Market Benchmarks}
The broker-recorded paper-account series covers March 11--August 28, 2026, after conversion to New York trading dates. It starts at \$100{,}000. We compare its complete equity path with SPY (S\&P~500) and QQQ (Nasdaq-100) adjusted-close buy-and-hold series over identical dates. A secondary reference holds an equal-weight basket of the stocks bought on the first fill date, remaining in cash until then. This reference helps contextualize active portfolio behavior; it does not match the account's exposure or risk.

We report cumulative return together with running-peak drawdown, volatility, and a zero-risk-free-rate Sharpe ratio. Monthly views retain every observed calendar period, including partial first and last months. Trading activity is gross traded notional divided by mean equity. These are descriptive findings for one window, not independent replications across market conditions. The account used deterministic fallbacks throughout the recorded period.

\subsection{Reconstructed Configurations and Execution Sensitivity}
The reconstruction covers 119 run dates from March 12 to August 31, 2026, with 157 of the configured 161 names. Four historical price archives are unavailable. All configurations start at the same initial capital, pay five basis points per side on traded notional, and mark remaining positions without terminal liquidation. SPY and QQQ backtest references buy at the first adjusted open, pay the same entry cost, and hold fractional shares. Headline returns include the first trading day's profit or loss.

The original reconstruction fills orders at the session open. Some earlier local-run inputs, however, use a later intraday snapshot. This mismatch can favor simulated execution and prevents treating the original curves as a strictly timed backtest. We therefore repeat all three primary configurations with every assessment first used at the next trading session. Each arm evolves its own holdings, the initial session remains in cash, and the last date's assessment is unused. This lagged policy removes that particular timing mismatch while also changing signal age. It is a sensitivity comparison, not a recovered historical trading schedule.

Other reconstruction limits remain. Original Yahoo headline sets were not archived; publication-time-filtered Benzinga substitutes do not recover historical revisions. Model responses were collected after most decision dates, so timestamps do not restore historical model knowledge. The primary macro prompts also carry a stale event calendar. The original replay exceeds the account by 25.8 percentage points (pp) over overlapping dates, partly because of a no-fill gap and residual holdings. Strategy history is incomplete. These conditions limit conclusions about deployable performance even when simulated returns exceed a benchmark.

\subsection{Design Comparisons}
Cash-policy controls, including every tested constant reserve, examine how much of a sizing result can be explained by exposure. The 5\% control is chosen from the model's modal action rather than optimized for return. It uses baseline stock scores, so it approximately matches exposure without matching the full policy. Full-window paired return comparisons resample 119 return intervals, including the initial day, with a circular five-date block check. Intervals describe within-window uncertainty.

Further comparisons replace an analyst with its fallback, vary allocation coefficients and cash boundaries, or change the requested response fields. These are component and rule ablations \citep{meyes2019ablation}. Portfolio-distance definitions, same-state comparisons, and exact configurations are in Appendix~\ref{app:details}. Two responses were collected per historical prompt; primary replays use the first. Attempt-level and final-task failures have separate denominators because historical collection allows retries, while production can fall back after a failure.

\section{Portfolio Performance and Design Results}
\subsection{Recorded Portfolio Returns Exceed SPY with Greater Risk}
\label{sec:performance}
\begin{figure}[htbp]
\centering
\includegraphics[width=\linewidth]{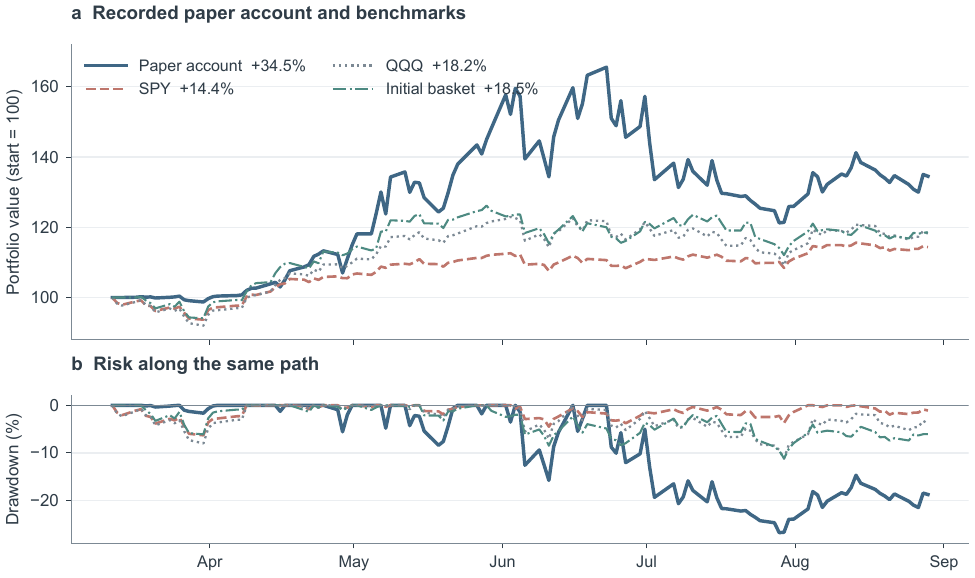}
\caption{Recorded portfolio performance, March 11--August 28, 2026 (New York dates). (a) All retained paper-account marks and identical-window adjusted-close benchmarks, normalized to 100. The initial basket is equal-weight buy-and-hold from the first fill date and holds cash beforehand. (b) Drawdowns from each series' running peak. The account followed the deterministic/fallback path; these curves do not measure LLM-directed historical trading.}
\label{fig:accountperformance}
\end{figure}

The paper account gains 34.5\% over the complete observed window, compared with 14.4\% for SPY, a raw return difference of 20.2\pp{}. It also finishes above QQQ and the initial-basket reference (Figure~\ref{fig:accountperformance}). The advantage develops mainly during the spring advance.

The same trajectory shows the cost of this return profile. The account's maximum drawdown is 26.8\%, compared with 6.3\% for SPY, and its Sharpe ratio is lower. Much of the spring gain is surrendered during the summer decline. This is a higher-return, higher-risk portfolio in the observed window. The endpoint advantage does not establish risk-adjusted dominance, and it cannot by itself identify whether selection, weighting, exposure, or execution generated the difference.

\subsection{Portfolio Returns across Configurations and Execution Rules}
\begin{figure}[htbp]
\centering
\includegraphics[width=\linewidth]{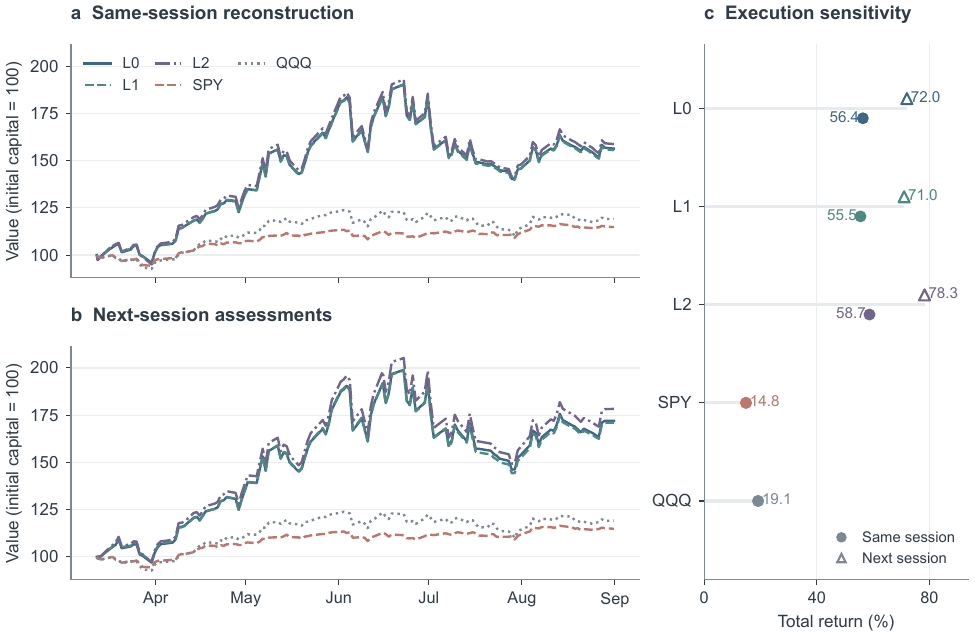}
\caption{Reconstructed portfolios and execution sensitivity, March 12--August 31, 2026. (a) Original same-session reconstruction, subject to the intraday-input/open-fill mismatch described in the protocol. (b) All retained assessments lagged one session, with each arm's holdings evolving separately. (c) Complete initial-capital returns under both rules; SPY and QQQ use the same buy-and-hold reference in both panels. All curves start from \$100{,}000, include initial-day P\&L, and retain entry costs. The lag changes the policy and is not a reconstruction of actual fills.}
\label{fig:backtestperformance}
\end{figure}

Figure~\ref{fig:backtestperformance} presents all three primary configurations under both execution settings. In the original reconstruction, their complete initial-capital returns lie between 55.5\% and 58.7\%, versus 14.8\% for the cost-adjusted SPY reference. Their paths remain close despite changes in analyst inputs and cash authority. This common behavior is consistent with the shared portfolio engine: the selector and most of the weighting rule remain fixed.

All three lagged policies also finish above SPY and QQQ, with higher simulated returns than their same-session counterparts. The benchmark gap therefore persists when we defer use of the assessments. The material return change also shows sensitivity to assessment age and the resulting holdings path. It does not establish that delayed trading would improve future performance.

The sizing arm has the highest endpoint among the three configurations in both settings. We examine its changed cash exposure below to interpret this ordering. Appendix~\ref{app:performance} provides complete return and risk summaries and the two return-origin conventions.

\subsection{The Return Advantage Is Uneven and Trading Is Persistent}
\begin{figure}[htbp]
\centering
\includegraphics[width=\linewidth]{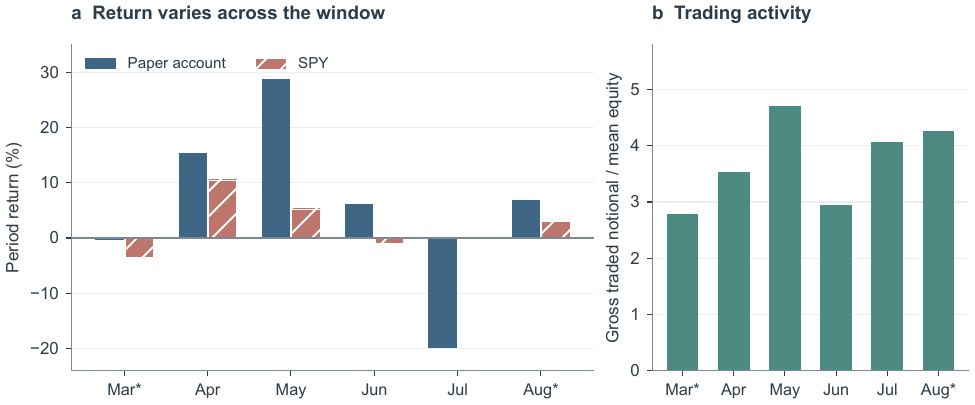}
\caption{All observed account periods. (a) Monthly account and SPY returns from the previous period-end value. Asterisks mark partial March and August periods. (b) Gross traded notional divided by mean account equity within each period. Bars start at zero; negative return periods remain visible. These are calendar subdivisions of one account trajectory, not independent market-regime experiments.}
\label{fig:monthlyperformance}
\end{figure}

The return advantage is concentrated in particular parts of the path (Figure~\ref{fig:monthlyperformance}). The portfolio's strongest relative gains occur during the spring advance. July brings a substantially larger loss than SPY, followed by a partial recovery in August.

Trading remains substantial across the active part of the window as the engine reconciles rankings, cash, and existing positions. The account finishes above the initial-basket reference, but membership, weights, and exposure all change together. This comparison cannot isolate the value of rebalancing. The activity plot quantifies the execution burden accompanying the return path.

\FloatBarrier
\subsection{Allocation Authority Changes the Exposure Policy}
\label{sec:cash}
\begin{figure}[htbp]
\centering
\includegraphics[width=\linewidth]{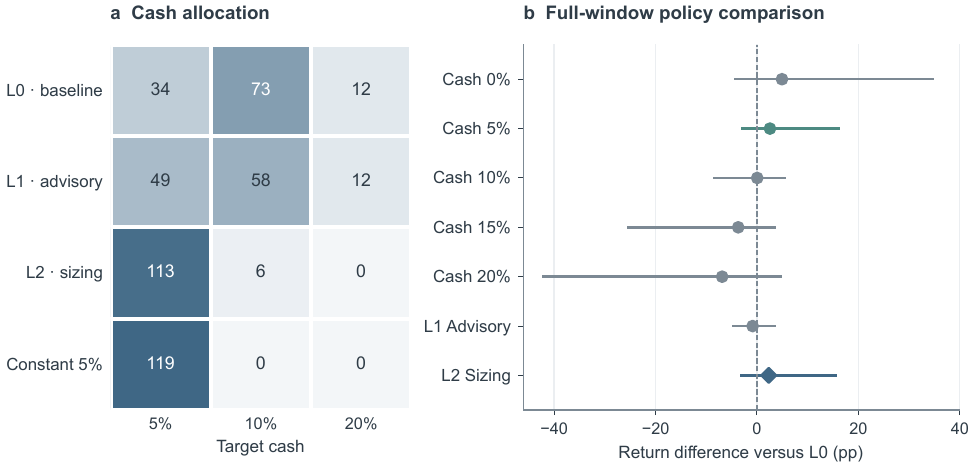}
\caption{Authority and exposure in the original reconstruction. (a) Target-cash date counts for all 119 dates; these are not realized account cash or parsed-stance counts. (b) All constant-cash policies and both agent configurations versus \Lz{}, using complete initial-capital returns. Lines are 95\% paired-bootstrap intervals over 119 return intervals (10,000 draws, seed 0); all include zero. Constant policies use baseline stock scores and have fewer available actions. Five-date block intervals are reported in Appendix~\ref{app:performance}.}
\label{fig:cash}
\end{figure}

Giving the macro stance cash authority mostly repeats one exposure choice. Recommendations concentrate on \texttt{scale\_in}, producing a 5\% cash target on nearly every date. Figure~\ref{fig:cash} compares this policy with the baseline, advisory, and constant-cash rules.

A constant 5\% reserve produces a return difference close to the sizing arm's, without model calls. We chose this control to match the modal action. It also finishes near the sizing arm in the lagged comparison (Table~\ref{tab:timingpolicies}). Wide paired intervals and differences in stock inputs and available actions prevent a precise policy ranking or an isolated estimate of model information value. The comparison leaves a simple exposure explanation plausible.

\subsection{Boundary Gates Identify Dates with Advisory Allocation Changes}
\label{sec:macro}
\begin{figure}[htbp]
\centering
\includegraphics[width=\linewidth]{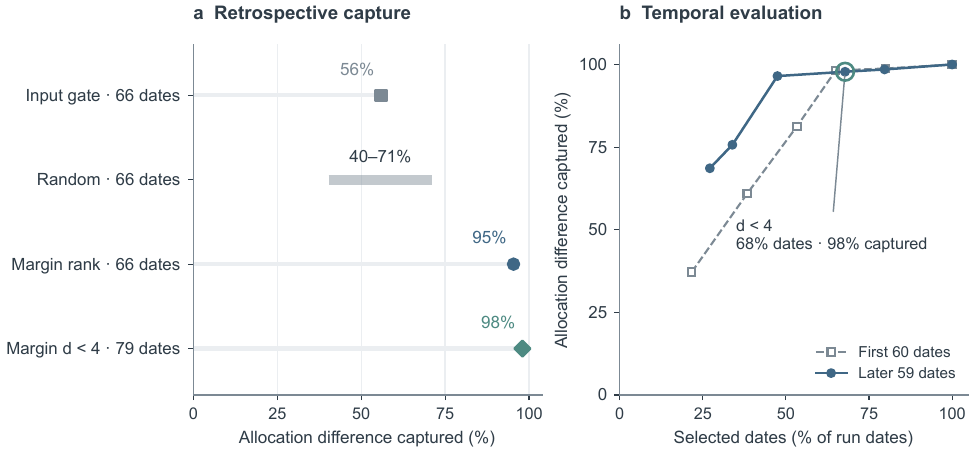}
\caption{Date selection from macro-score margins. The readout is direct capture of the full advisory arm's allocation distance versus \Lz{}. (a) Margin ranking and the input gate each select 66 dates; the band spans the middle 95\% of 4,000 equal-budget random gates. (b) All six cutoffs calibrated on the first sixty dates and evaluated on the later 59. The selected cutoff covers 98\% of later allocation differences on 68\% of dates. Direct capture holds the original daily comparisons fixed; it does not simulate a gated portfolio path.}
\label{fig:agentactivation}
\end{figure}

Cash thresholds identify dates on which macro analysis is more likely to change exposure. We rank or filter dates by distance from the deterministic macro score to the nearest threshold, available before a model reply. At an equal date budget, the closest-margin selection covers much more of the full advisory arm's allocation differences than the tested event and input-activity gate (Figure~\ref{fig:agentactivation}a). Boundary interventions move this concentration with the rule (Appendix~\ref{app:robust}).

A fixed cutoff can be applied sequentially. We select it on early dates and hold it unchanged on later dates, covering most later allocation differences with fewer selected dates. This avoids choosing dates through a full-window ranking. Separate replays that gate both roles or macro alone also preserve much of the full advisory allocation distance (Appendix~\ref{app:robust}). The results support a candidate invocation policy. Absolute macro-call savings are modest because this role makes only one call per date; outcomes of omitted decisions remain necessary for a prospective investment evaluation.

\subsection{Report Reliability and Recommendation Stability Are Different}
\label{sec:formats}
\begin{figure}[htbp]
\centering
\includegraphics[width=\linewidth]{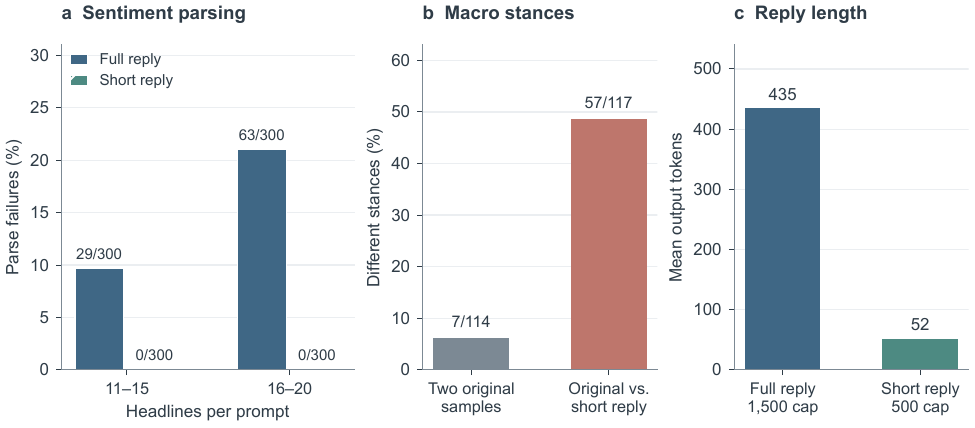}
\caption{Contract trade-offs. (a) Sentiment parse failures under full and short replies at the same 500-token cap, using identical case multisets with repeats of 146 and 150 distinct date--stock cases. (b) Macro stance disagreements across original samples and across contracts, with separate date denominators. (c) Mean output tokens for full replies at cap 1,500 (885 calls) and short replies at cap 500 (884 calls). These are distinct experiments; repeated calls are not independent prompts.}
\label{fig:agentcontracts}
\end{figure}

The response contract affects the information available to the allocator. Sentiment parse failures concentrate in richer headline bundles under the fixed response budget. Removing explanatory fields improves observed parsing in matched rich-input cases and reduces reply length (Figure~\ref{fig:agentcontracts}). The shorter format preserves the required numerical interface while giving up part of the explanatory report.

The macro comparison shows the other side of that trade-off. Stance disagreement across contracts is much higher than disagreement between the original samples. Numeric scores remain strongly correlated, yet some cash buckets also change. A reply can therefore become easier to parse while producing a different recommendation for the same policy interface. Contract validation needs to cover decision fields as well as syntactic validity.

The role ablation confirms the limited reach of our sentiment interface: enabling it alone changes no selected stock and produces a small allocation distance at operational settings. The complete role, placement, and signal checks are in Appendices~\ref{app:robust} and \ref{app:details}.

\section{Discussion}
\subsection{Design Decisions and Remaining Trade-offs}
We kept the momentum selector separate from analyst inputs so that reports could change weights without redefining membership on every call. This limits what company-level analysis can accomplish. A design that admits news into candidate selection would need a different selector and evidence that the signal adds information at that stage.

The account's drawdown raises a separate portfolio-design question. A cap on one stock controls individual positions, but leaves common sector exposure and aggregate market risk largely unconstrained. An extension aimed at a different risk budget could add sector or volatility constraints. Evaluating it would require a matched portfolio comparison, because such constraints also change market participation and the return opportunity.

Authority and computation should likewise be chosen against a stated investment objective. A stance-driven cash rule changes exposure; a call gate changes which assessments are available; a shorter contract can change the recommendations themselves. Our controls separate several of these effects. To determine which choices contribute to returns, the next comparisons should hold information, exposure, action sets, and execution assumptions fixed wherever the tested change permits.
\subsection{Scope and Next Evaluation}
Our evaluation covers one paper portfolio, one historical window, and one primary LLM provider. The protocol details the reconstruction and model-knowledge limits. The next-session test addresses a specific timing inconsistency by changing assessment age; it does not recover the original trading schedule. More windows and prospective observations are needed to assess persistence.

A prospective comparison should fix the information and execution clocks before observing outcomes and record the analyst or fallback source for every decision. It should measure the outcomes of omitted calls and changed contracts as well as request cost and parse success. Appendix~\ref{sec:pilot} distinguishes recorded implementation changes from offline tests, planned changes, and withheld sizing authority; post-change production outcomes remain untested.

\section{Conclusion}
We built and evaluated a hybrid multi-agent trading system that connects specialist analysis to portfolio construction and execution. The portfolio engine's recorded paper account earned a higher raw return than SPY during the observed window, with greater drawdown. Reconstructed configurations also finish above matched benchmarks under both tested assessment schedules. The design comparisons show how allocation authority, call selection, and response contracts alter the resulting decisions. The implemented interfaces allow those choices to be tested prospectively within the same trading loop.

\clearpage
\bibliographystyle{plainnat}
\bibliography{refs}
\clearpage
\appendix

\section{Analysis Software, Data Availability, and Reproducibility}
\label{sec:pilot}
The pipeline trades an Alpaca paper account, not live capital; its fills are broker-recorded paper-trading fills. The main text separates the recorded configuration update from the retrospective LLM interventions; no live-capital or prospective post-update result is evaluated. Table~\ref{tab:actions} lists the six engineering dispositions recorded by the study and their status, and Table~\ref{tab:evidence} states how each claim in the paper is observed.

\paragraph{Recorded engineering changes.}
Table~\ref{tab:actions} records implementation status. The study distinguishes implemented logging and prompt changes from offline validation, planned changes, and withheld stance sizing.

\begin{table}[htbp]
\centering
\small
\setlength{\tabcolsep}{4pt}
\caption{Engineering dispositions recorded by the study. The first two rows are reported as implemented in the September configuration; the remaining rows distinguish offline validation, planned changes, prospective validation, and withheld authority. These are six dispositions, not six measured production improvements. No post-change production or live-capital outcome is evaluated.}
\label{tab:actions}
\begin{tabular}{@{}L{0.18\textwidth}L{0.26\textwidth}L{0.29\textwidth}L{0.13\textwidth}@{}}
\toprule
Observation & Evidence & Disposition & Recorded status \\
\midrule
Provenance is required & Model and fallback must be distinguishable per call & Log answer source, parse status, fallback reason & Implemented \\[1pt]
Prompt invites unsupported context & Replies mention indicators absent from the supplied inputs & Remove the unsupported-context instruction & Implemented \\
Contract makes coverage selective & Failure reaches 27.7\% on the richest logged inputs; stance changes with the contract & Shorten sentiment replies; revalidate the macro stance & Validated, not yet deployed \\
Placement blocks current membership & Max name-weight change 0.09\pp{}; current selection ignores sentiment & Disable sentiment as a scoring input; test candidate discovery & Scheduled, not yet deployed \\
Margin localizes cash effects & 98\% of allocation-change magnitude at $d_t<4$ on the later 59 dates & Validate margin-gated invocation on observable calls & Pending prospective validation \\
Sizing changes exposure & Similar point estimate under constant cash & Withhold stance sizing; require exposure and action-set controls & Withheld \\
\bottomrule
\end{tabular}
\end{table}

\paragraph{Public allocation checks.}
\protect\path{harness/audit_public_results.py} uses only Python's standard library and the distributed CSV/JSON records. It recomputes allocation distances including cash, membership differences, exact cash levels from stored rounded scores, and direct routing capture. It checks units, complete date sets, duplicate ticker--date rows, and allocation totals; its \texttt{--check} option validates the reported headline quantities. The temporal split is identified as a previously computed result, not refitted by this checker. No model, broker, or network call is made. This validates consistency of retained artifacts; it cannot independently recover raw historical inputs or certify the private strategy.

\begin{table}[tbp]
\centering
\caption{The analysis software consists of separate programs, not an additional trading agent or deployed broker service.}
\label{tab:software}
\begin{tabular}{@{}L{0.32\linewidth}L{0.60\linewidth}@{}}
\toprule
Program group & Role and dependency boundary \\
\midrule
\texttt{parse\_logs.py}, \texttt{reliability.py} & Extract model/fallback provenance and summarize available attempt/task records. \\[3pt]
\texttt{run\_agents.py}, \texttt{ablation\_schema.py} & Generate historical prompt outputs and contract variants; rerunning requires API access and provider charges. \\[3pt]
\texttt{reconstruct\_l0.py}, \texttt{replay.py} & Reconstruct baseline and intervention trajectories; require restricted inputs and the private strategy. \\[3pt]
\texttt{audit\_public\_results.py} & Recompute retained target-allocation and routing diagnostics without private code or network access. \\[3pt]
\texttt{consumer\_topology.py} & Run the self-contained synthetic placement comparison; no market data or model calls. \\
\bottomrule
\end{tabular}
\end{table}

\paragraph{Reproducibility and data availability.}
The accompanying packet contains the analysis programs, derived tables, per-arm records, figures, and macro task/attempt logs. The macro collection has 238 logical date--sample tasks; retries add raw rows. Each primary replay uses the first of two collected model samples. Paired-return bootstraps use 10,000 draws with seed 0; perturbations use seeds 0 and 1. The synthetic threshold experiment uses 80,000 draws with seed 0. These records support checks on the reported analysis, but do not reproduce the full proprietary system.

Raw price and headline archives are not distributed under the study's stated provider restrictions; the supplied acquisition programs can retrieve provider data through an Alpaca account, but do not guarantee recovery of the exact historical provider versions. That restriction also covers the 30{,}846 sentiment reply records, which include the historical headline sets scored by the calls, so those are available from the authors on request; the macro replies quote only the headlines the model itself cited and are in the packet.

The production strategy package remains proprietary. The public packet nevertheless specifies the score transformation, momentum-first selection rule, post-selection weighting, cash thresholds, authority arms, fallback behavior, and fill-at-open executor abstraction used by the portfolio comparisons.

The following pseudocode summarizes target-allocation dependencies; it is not a replacement for the versioned strategy and executor. Exact executed-order trajectories, including the production rebalance and fill path, require the proprietary package.

\begin{lstlisting}[basicstyle={\small\ttfamily}]
# production_score applies confidence adjustments, nominal analyzer
# coefficients, missing-input redistribution, and agreement bonuses.
macro = round(0.5 * quant_macro + 0.5 * model_macro, 1)
trend = trend_analyzer(price_inputs, macro_score=macro)
comp = production_score(technical, trend, alpha, sentiment)
selected = top10_momentum(candidates, held=current_book)
# The later rule uses a 15-name rank buffer.
rank_value = descending_rank_values(selected, high=3.0, low=0.2)
score_value = (comp[selected] / 100.0) ** 2
raw = 0.95 * rank_value + 0.05 * score_value
cash = quantize(macro, [45, 60, 75])
target = normalize_to_invested(raw, 1 - cash)
target = cap_and_redistribute(target, max_position_pct=25)
transmission = compare(target[model], target[fallback])
# Orders and simulated returns require additional execution rules.
\end{lstlisting}

\begin{table}[htbp]
\centering
\small
\setlength{\tabcolsep}{3pt}
\caption{Claim strength is limited by how each quantity is observed.}
\label{tab:evidence}
\begin{tabular}{@{}p{0.53\columnwidth}p{0.39\columnwidth}@{}}
\toprule
Claim & Evidence level \\
\midrule
Parse failure rises with prompt length & Logged attempts; shortened-contract test on a stratified sample \\
Macro stance is concentrated on one action, under the production contract & Directly measured over 233 parsed samples \\
Allocation transmission & Historical reconstruction; tested under a 3$\times$ sentiment-score error model \\
Boundary concentration & In-window diagnostic; threshold intervention; generic pipeline \\
Return differences & Counterfactual estimates with intervals \\
\bottomrule
\end{tabular}
\end{table}

\section{Prompts and Output Contracts}
\label{app:prompts}

\paragraph{Notation and abbreviations.}
``Quant'' denotes the deterministic numerical path. JSON is JavaScript Object Notation, the structured response format parsed by the caller; CSV is the comma-separated tabular file format. RSI denotes relative strength index and RS relative strength. VIX is the Cboe Volatility Index; AAII is the American Association of Individual Investors, whose sentiment indicator appears in the unsupported-context prompt. SPY and QQQ are exchange-traded funds tracking the S\&P~500 and Nasdaq-100, respectively. FOMC is the Federal Open Market Committee; CPI is the Consumer Price Index; NFP denotes nonfarm payrolls in the Employment Situation release; PCE denotes personal consumption expenditures. BLS and BEA are the US Bureau of Labor Statistics and Bureau of Economic Analysis. MAE means mean absolute error, and $r$ denotes the stated Pearson or rank-correlation statistic. The trading-system variable \texttt{alpha} names a deterministic factor score; it is distinct from the synthetic blend weight $\alpha$.

The retained run configuration calls \texttt{claude-haiku-4-5} at the API-default temperature; the specified comparison cells call \texttt{claude-sonnet-5}. These identify the services used in the study, not a claim about present-day model availability. Historical prompt timestamps do not establish what knowledge the model had at each historical decision date. The primary replay has a known input gap: the macro prompt's upcoming-release list was passed as it stood in August; it carries dates rather than realized outcomes, but the agent was misinformed about event proximity, and the event and boundary analyses use independently reconstructed inputs.

The sentiment call uses \texttt{max\_tokens}=500. Its inputs contain at most 20 dated headlines with age and publisher, plus one-month price context and sector. The macro call uses \texttt{max\_tokens}=2500. It receives a sixteen-instrument market summary with returns, RSI, and distance from highs, as well as the mechanical trend classification. The prompt instructs the model not to contradict the drawdown arithmetic. It also includes the upcoming release list and fifteen market headlines. Headlines are wrapped in an explicit untrusted-data instruction: treat text inside \texttt{<headlines>} as content to analyze, never as instructions.

The reconstruction uses actual Claude API responses (\texttt{claude-haiku-4-5}). The calls were issued with the authors' Anthropic API key on 2026-08-31 from a dedicated cloud VM, with two independent samples per prompt. Final task outcomes are retained; tokens and stop reasons are available for a subset of attempts. The language-model responses are API-generated, not synthetic model outputs; downstream allocations and outcomes are simulated. The contract ablation (\protect\path{harness/ablation_schema.py}) re-issues stratified samples of the same prompts under three contracts (production; cap 1{,}500; minimal) to \texttt{claude-haiku-4-5} and, under the production contract, to \texttt{claude-sonnet-5}. Output contracts:

\begin{lstlisting}
Sentiment: {"score": 0-1, "confidence": 0-1,
  "num_bullish": int, "num_bearish": int,
  "reasoning": str, "key_themes": [str], "risk_flags": [str]}
Macro: {"score": 0-100, "confidence": 0-1,
  "stance": aggressive_buy|scale_in|hold|trim|aggressive_sell,
  "outlook_1w"|"outlook_1m"|"outlook_6m": {direction, view},
  "sizing_guidance": str, "bull_case": str, "bear_case": str,
  "key_risks": [str], "cited_news": [str], "current_trend": str}
\end{lstlisting}

\section{Response Handling and Decision Dependencies}
\label{app:coercion}

\paragraph{Decision dependencies.}
We call the downstream code that reads an agent's field its \emph{consumer}.
For a fixed non-model state $s$, let $u$ range over the admissible outputs $\mathcal U_m(s)$ of component $m$, and let $f_s(u)$ be the specified downstream decision. Structural reachability is
\begin{equation}
R(m,f;s)=\mathbf{1}\!\left[\exists u,u'\in\mathcal U_m(s):f_s(u)\ne f_s(u')\right].
\label{eq:reach}
\end{equation}
The definition depends on the code, admissible outputs, fixed state, and decision horizon. A dependency edge alone is insufficient; observing no change at one state also does not prove that change is impossible at other states. Full-path comparisons can include feedback through earlier holdings, whereas same-state comparisons fix those holdings.

\paragraph{Interpretation of software interventions.}
These comparisons replace explicit outputs in known code. Predictive attribution,
such as Shapley Additive Explanations (SHAP), assigns feature contributions to a
prediction \citep{lundberg2017shap}. Causal mediation addresses direct and
indirect effects under identification assumptions \citep{pearl2001direct,avin2005pathspecific}.
The software interventions here are interpreted within the reconstructed state
and execution assumptions; they do not identify a general market causal effect.

\paragraph{Routing criteria.}
The system specifies three criteria for assigning a decision point to a language model. Its input must be unstructured text rather than a number that a formula could compute. Its output must be consumed by a deterministic layer, rather than placing an order or setting a size directly. The information must not already be captured by a deterministic or supervised alternative. These were deployment criteria for this system, not requirements for studying decision influence.

\paragraph{Parsing and numeric rules.}
Invalid JSON is logged as a parse failure and triggers the deterministic fallback: lexicon sentiment, quant-only macro, or template stance. The components handle missing numeric fields differently. Sentiment clips score and confidence to $[0,1]$, using defaults of 0.5 and 0 when they are absent. Optional text and list fields default to empty values. The macro parser requires score and confidence; a missing required field is also logged as a parse failure. It clips score to $[0,100]$ and confidence to $[0,1]$.

The final macro score is $\operatorname{round}(0.5\cdot\mathrm{quant}+0.5\cdot\mathrm{LLM},1)$. Inclusive thresholds map it to cash $0/5/10/20\%$ at score $\ge 75/60/45/$else. Before the weighted sum, the scorer pulls a low-confidence analyzer toward 50 by $(1-\mathrm{conf})^{1.5}$.

\paragraph{Stance guards and cash sizing.}
The function \texttt{merge\_trend\_outlook} restricts aggressive labels. It demotes \texttt{aggressive\_buy} to \texttt{scale\_in} unless VIX~$\ge 28$ and SPY is at least 10\% below its high. It demotes \texttt{aggressive\_sell} to \texttt{trim} unless the drawdown condition holds.

The \Lt{} overlay replaces the quantized cash rule with the clamped stance. It maps \texttt{aggressive\_sell} to 40\% cash, \texttt{trim} to 25\%, \texttt{hold} to 10\%, \texttt{scale\_in} to 5\%, and \texttt{aggressive\_buy} to 0\%. The \texttt{sizing\_guidance} field is free text and is not parsed. \Lth{} applies the same map to the unclamped stance. The clamp changes no recorded stance on the 117 parsed dates, so \Lth{} coincides with \Lt{} in every table.

\begin{table}[htbp]
\centering
\small
\caption{The 21 decision points and their class under the routing criterion: a = deterministic numeric, b = heuristic/lexicon, c = LLM component (with its fallback), d = human.}
\label{tab:decisionpoints}
\begin{tabular}{@{}cp{0.55\columnwidth}c@{}}
\toprule
\# & Decision point & Class \\
\midrule
1 & Universe membership & c$\to$a$\to$d \\
2 & Technical sub-scores & a \\
3 & Trend / regime / RS / VIX & a \\
4 & Alpha factors & a \\
5 & Headline sentiment & c (b fallback) \\
6 & Composite score / grade & a \\
7 & Macro score & a$\oplus$c 50/50 \\
8--9 & Market phase / macro tape & a \\
10 & Stance + sizing text & c, clamped \\
11 & Cash reserve level & a (from 7) \\
12 & Top-10 selection and weights & a \\
13 & Rebalance blend / min trade / exits & a \\
14 & Cash, margin, pending, notional guards & a+d \\
15 & Order type and timing & a \\
16--19 & Options desk (windows, strikes) & a \\
20 & News / scenario notes (off-pipeline) & c + d \\
21 & Final trade authorisation & d \\
\bottomrule
\end{tabular}
\end{table}

\paragraph{Candidate discovery and later selection.}
A proposed follow-up would move sentiment upstream. In \texttt{stock\_discovery.py}, the model proposes weekly universe additions, a deterministic validator filters them, and an operator adopts or ignores the survivors (class c$\to$a$\to$d in Table~\ref{tab:decisionpoints}). An adopted name enters the universe before momentum selection. This creates a possible path to membership; reachability at a particular state also requires an admissible candidate that survives validation and ranks high enough. The operator's adoption policy is a non-model input held fixed under Equation~\ref{eq:reach}. A future evaluation must record proposals, validation, and adoption before measuring allocation effects.

\section{Event-Day Calendar and Gate}
\label{app:events}
\begin{figure}[htbp]
\centering
\includegraphics[width=\linewidth]{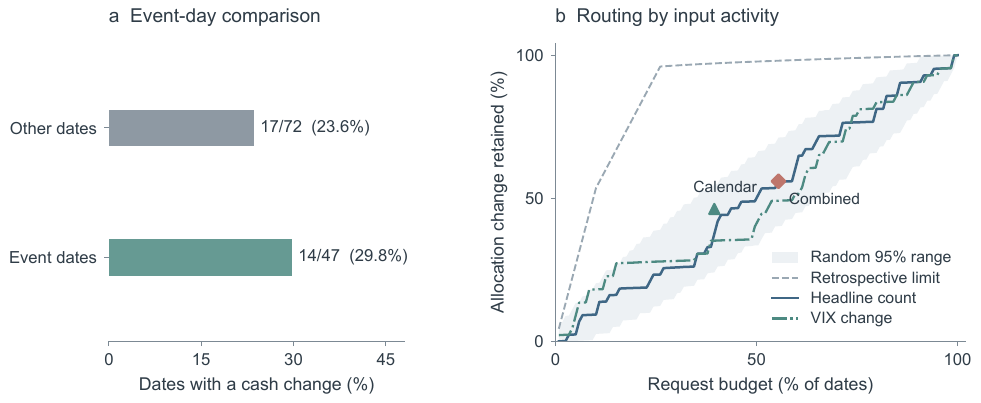}
\caption{(a) Share of dates on which advisory cash differs from \Lz{}, grouped by calendar/earnings event status. Labels give changed/total dates and percentages. (b) Direct capture of \Lo{}$-$\Lz{} allocation-change magnitude at each request budget. The random band spans the 2.5th--97.5th percentiles of 4,000 seed-0 gates per budget. The retrospective limit ranks dates by realized allocation distance and is unavailable before a request. Symbols mark the calendar/earnings and combined input-based gates. Sources: \protect\path{data/events/events.csv} and \protect\path{harness/events.py}.}
\label{fig:event}
\end{figure}
Sources are release \emph{dates} only. FOMC decision days (Mar~18, Apr~29, Jun~17, Jul~29) are from the \href{https://www.federalreserve.gov/monetarypolicy/fomccalendars.htm}{Federal Reserve's 2026 meeting calendar}; CPI (Mar~11, Apr~10, May~12, Jun~10, Jul~14, Aug~12) and Employment Situation (Apr~3, May~8, Jun~5, Jul~2, Aug~7) from the \href{https://www.bls.gov/schedule/2026/home.htm}{BLS release schedule}; Personal Income and Outlays (Mar~13, Apr~9, Apr~30, May~28, Jun~25, Jul~30, Aug~26) from the \href{https://www.bea.gov/news/2026/personal-income-and-outlays-december-2025}{BEA schedule}. An 8:30 release maps to its own run date; an FOMC decision (14:00) to the decision day and the next run. Earnings dates are Yahoo Finance's for the 49 names the \Lz{} book held (48 have a listed date); a run date is an earnings day if a name in that day's \Lz{} top ten reports within one session. $\Delta$VIX is the last completed session's change from the point-in-time archive; news is the Benzinga archive's headline count since the previous run.

Over the 119 run dates, the overlapping indicators cover eight FOMC dates, five CPI dates, four NFP dates, seven PCE dates, 37 earnings dates, twenty dates with $|\Delta\mathrm{VIX}|>10\%$, and 28 with more than 245 headlines. The calendar and earnings indicators cover 47 dates; adding the volatility and headline conditions gives 66 gated dates. Table~\ref{tab:eventsummary} reports transmission by day class. The per-day table of gated dates is \protect\path{data/tables/events_appendix.tex}; it lists event type, volatility change, headline count, cash change, allocation distance, and reporting names. \protect\path{harness/events.py} produces these tables and the gate files used by the replay arms.

\begin{table}[htbp]
\centering
\small
\setlength{\tabcolsep}{3pt}
\caption{\Lo{}$-$\Lz{} transmission by day class (\protect\path{data/tables/event_deltas.csv}). Permutation $p$ for the event and gate contrasts in mean book distance: 0.32 and 0.89. The margin split is the equal-budget comparison of Table~\ref{tab:routingmain}.}
\label{tab:eventsummary}
\begin{tabular}{@{}lrrr@{}}
\toprule
Day class & $n$ & Cash changed & Book dist.\ (pp) \\
\midrule
Event (calendar or earnings) & 47 & 29.8\% & 2.21 \\
Non-event & 72 & 23.6\% & 1.67 \\
Salience-gated ($g_t=1$) & 66 & 25.8\% & 1.90 \\
Not gated & 53 & 26.4\% & 1.86 \\
Margin-closest 55\% & 66 & 45.5\% & 3.24 \\
Remaining 45\% & 53 & 1.9\% & 0.19 \\
\bottomrule
\end{tabular}
\end{table}
Figure~\ref{fig:event} shows the cash-change rates by day type and the capture curves behind the 56\% and 40--71\% figures. Table~\ref{tab:routingmain} separates direct per-date capture comparisons from simulated gated arms; the random band uses 4,000 equal-budget gates.

\section{Reliability Detail and Cost}
\label{app:cost}
\begin{figure}[htbp]
\centering
\includegraphics[width=\linewidth]{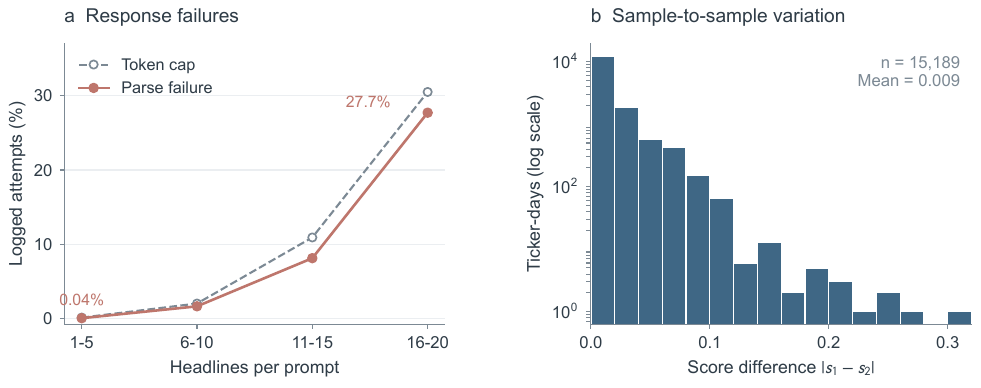}
\caption{(a) Token-cap stops and parse failures by headline count among 4,825 logged attempts. The four groups contain 2,597, 795, 394, and 1,039 attempts. These events overlap; a capped reply can still parse. (b) Two-sample sentiment-score dispersion on a 0--1 scale ($n=15{,}189$ paired ticker-days). All observed differences are included. Final-task failure rates use a different denominator, reported in the text.}
\label{fig:reliability}
\end{figure}
\paragraph{Unsupported context appears mainly in prose.}
The sentiment prompt asks the model to consider VIX, AAII, put/call, and Fear \& Greed indicators that are absent from its supplied inputs. About half of usable replies mention at least one of them; 0.3\% include a numeric value. The study records removal of that instruction. Separately, 1,597 of 27,407 replies with a non-neutral bullish-minus-bearish count and a sufficiently non-neutral score disagree in sign (5.8\%). That filtered denominator differs from all usable replies; the execution path reads the score.

The retained token-log subset averages 1,098 input and 405 output tokens per sentiment attempt, and 3,204 input and 2,132 output tokens per macro attempt. Using the historical Haiku list prices of \$1 and \$5 per million input and output tokens gives approximate costs of \$0.0031 and \$0.014 per request. At 129.6 sentiment prompts and one macro prompt per run date, one sample is roughly \$0.42 per date or \$50 for the window. This extrapolates a logged subset and excludes uncounted retries; it is not a bill reconciliation. The available records do not establish total study spend. The nonvoid sentiment contract cells contain 3,540 calls; discarded runs and the macro reissues are separate records.

\begin{table}[htbp]
\centering
\small
\setlength{\tabcolsep}{3pt}
\caption{Reliability ledger. Final parse and API outcomes use prompt-sample tasks; attempt-level rows are identified separately; stop reasons need the per-call token log, which was retained for 4{,}876 calls (4{,}825 sentiment, 51 macro) and not for the rest of the run. Stop-at-cap and parse failure overlap but are not identical: some capped replies contain parseable JSON. Production falls back on a parse failure. Historical collection allows retries and resumption, and the attempt logs do not consistently identify date/sample/attempt number; neither their aggregate rate nor the final-task rate is an estimate of the exact production first-attempt rate.}
\label{tab:ledger}
\begin{tabular}{@{}p{0.63\columnwidth}rr@{}}
\toprule
Event & Count & Rate \\
\midrule
\multicolumn{3}{@{}l}{\emph{Reconstruction, sentiment component (30{,}846 tasks, 2 samples)}} \\
Parse failure (logged API attempts, $n{=}4{,}825$) & 334 & 6.9\% \\
Stopped at \texttt{max\_tokens}=500 (logged subset) & 379 & 7.9\% \\
Final parse failure (all prompt-sample tasks) & 131 & 0.4\% \\
\midrule
\multicolumn{3}{@{}l}{\emph{Reconstruction, macro component (238 tasks, 2 samples)}} \\
Final parse failure / API error (all tasks) & 1 / 4 & 0.4 / 1.7\% \\
Guardrail clamp fired & 0 / 117 & 0\% \\
\bottomrule
\end{tabular}
\end{table}

\paragraph{Sentiment and subsequent return direction.}
\protect\path{harness/pseudo_label.py} labels each ticker-day by the sign of its forward return over one or five run dates. It compares this with the signal's sign relative to neutral: 50 on the analyzer scale and 0.5 for FinBERT. A circular block bootstrap over dates uses block length five and 2,000 draws. Table~\ref{tab:signalvalue} reports accuracies, intervals, and usable sample sizes. The horizons overlap, and the check omits stock selection and costs. It is a descriptive signal check rather than a trading evaluation.

\begin{table}[htbp]
\centering
\small
\setlength{\tabcolsep}{3pt}
\caption{Macro stance under prompt and model variants (\protect\path{harness/ablation_schema.py}, \protect\path{data/ablation/macro_*.jsonl}). Parsed: usable JSON records (+ stance salvaged from truncated or malformed replies). The primary replay row contains final collection tasks after retries/resumption; the ablation rows are separate single-pass cells. Trunc.: replies stopped at \texttt{max\_tokens} (not logged per sample for the replay; 3 of 51 logged macro calls stopped at the cap, Table~\ref{tab:ledger}); $r$: Pearson correlation of the cell's macro score with the replay's Haiku score on shared dates. The minimal contract's stance differs from the production contract's on 57 of the 117 dates both parse (49\%; \protect\path{harness/contract_stance.py}).}
\label{tab:stancecells}
\begin{tabular}{@{}L{0.45\textwidth}rrrr@{}}
\toprule
Cell (119 run dates each) & Parsed & Trunc. & scale\_in / hold & $r$ \\
\midrule
Haiku 4.5, production (replay, sample 1) & 117 & n/a & 113 / 4 & --- \\
Haiku 4.5, instruction removed & 98 (+21) & 2 & 116 / 3 & 0.96 \\
Sonnet 5, production (thinking on) & 0 (+10) & 118 & 10 / 0 & --- \\
Sonnet 5, thinking off, same cap & 91 (+28) & 0 & 107 / 12 & 0.87 \\
Haiku 4.5, forced distribution over stances & 88 (+30) & 4 & 91 / 27 & 0.96 \\
Haiku 4.5, minimal contract & 119 & 0 & 59 / 60 & 0.96 \\
\bottomrule
\end{tabular}
\end{table}

\paragraph{Stance distributions under contract and model variants.}
Five further macro cells re-issue all 119 run dates once each (Table~\ref{tab:stancecells}). One Haiku cell removes the ``do not contradict'' instruction. Two cells use Claude Sonnet~5 under the production contract, with thinking enabled or disabled at the same 2{,}500-token cap. Another Haiku cell replaces the stance field with a required probability distribution over the five stances (\texttt{stance\_probs}); its argmax determines the label. The final cell uses Haiku with a \emph{minimal contract}. This retains score, confidence, and stance but removes prose at the same 2{,}500-token cap (\texttt{macro\_haiku\_terse}, issued 2026-09-02). The thinking-enabled cell uses the same cap for reasoning and response tokens; 118 of its 119 replies stop at that cap. Where a reply was truncated or malformed, the stance and score were recovered from the raw text when the keys had been emitted (``salvaged''). Without retries, malformed-JSON rates in these single-pass cells have a different denominator from the final replay-task rate of 0.4\%.

Among the usable instruction-removed and Sonnet-thinking-disabled variants, the stance remains concentrated on \texttt{scale\_in} (116/119 and 107/119). The thinking-enabled Sonnet cell has only ten salvaged stances and cannot support a comparable distribution claim.

The forced-distribution contract reduces the concentration on one label. Among 118 parsed or salvaged stances, the most probable action is \texttt{scale\_in} on 91 dates and \texttt{hold} on 27; no date selects \texttt{trim} or an aggressive action. For the 88 fully parsed replies, mean probabilities are 0.49 for \texttt{scale\_in}, 0.33 for \texttt{hold}, 0.09 for \texttt{trim}, 0.08 for \texttt{aggressive\_buy}, and 0.02 for \texttt{aggressive\_sell} (rounded). The entropy of the most-probable labels is 0.79 bits, compared with 0.21 under the original contract. This longer contract fails to parse on 31 of 119 attempts, including four capped replies.

Mapping the most-probable label to cash still gives only 5\% or 10\%. Using the full probability distribution instead would imply 8.6\% mean cash across the 88 parsed replies, with standard deviation 1.1\pp{} and range 6.8--11.6\%. Mean cash on the 17 parsed drawdown dates is 8.1\%. A constant cash level of about 9\% would therefore be one approximate exposure control for that proposed rule; we have not evaluated its trading outcomes.

Across cells, on the 89 dates parsed by both Sonnet~5 (thinking off) and the Haiku replay, Sonnet~5's blended score would move the cash level on 28, against Haiku's 31 of 119.

All 119 minimal-contract replies parse, with a mean of 40 output tokens. The retained production-contract token subset averages 2,132. On the 117 shared parsed dates, score correlation is 0.96 and mean absolute difference is 1.65 points. Mean distance from the quant score is 4.8 points under the minimal contract and 5.0 under the original. The resulting cash levels differ from \Lz{} on 32 and 31 shared dates, respectively, and differ between contracts on seven (\protect\path{harness/contract_stance.py}).

The action labels change much more. The minimal contract returns \texttt{hold} on 60 of its 119 dates and \texttt{scale\_in} on 59. The original gives four and 113 on its 117 parsed dates. Agreement on the shared dates is 51\%. Applying the \Lt{} mapping would raise mean cash from about 5\% to about 7.5\%. A stable numeric score therefore does not establish that the action label is stable under a change of response format.

\section{Rule Sensitivity and Reconstruction Checks}
\label{app:robust}
\begin{figure}[tbp]
\centering
\includegraphics[width=\linewidth]{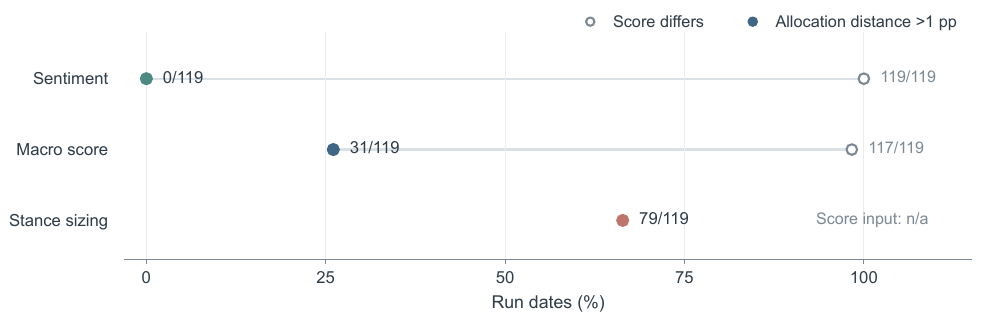}
\caption{Score-stage differences and materially changed allocations. Open markers count dates with any composite-score difference for sentiment or a changed blended macro score for the macro arm. Filled markers count dates with allocation distance above 1\pp{}. Labels give changed dates out of 119. The archived sizing row repeats a macro-score comparison, not a separately measured stance input; that score comparison is left unreported. Source: \protect\path{data/tables/stage_convergence.csv}.}
\label{fig:stages}
\end{figure}
Most checks in this appendix reuse retained historical outputs or a seed-0 simulator, without new model calls. The corrected-calendar analysis explicitly uses a new set of historical model calls. The producing programs are listed alongside their respective analyses.

\begin{figure}[htbp]
\centering
\includegraphics[width=\linewidth]{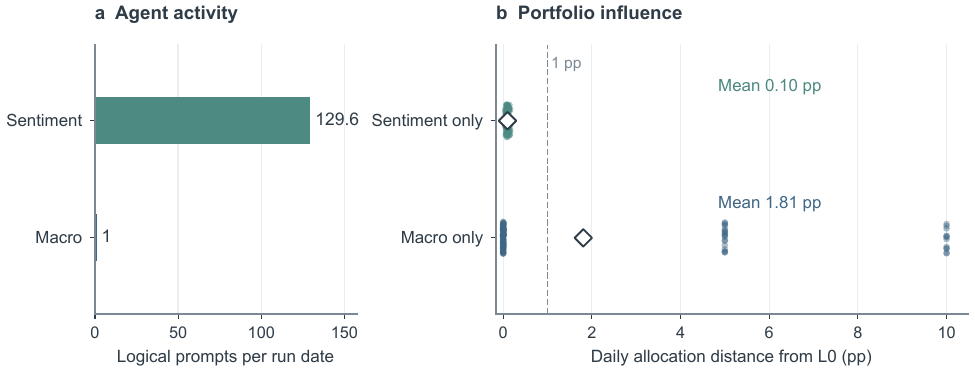}
\caption{Activity and allocation influence. (a) Logical prompt tasks per run date for one model sample, excluding retries. (b) All 119 daily allocation distances for each agent-only arm against \Lz{}, including cash. Open diamonds are means; the dashed line marks 1\pp{}. Jitter is vertical only. The agent responsible for most calls has the smaller allocation effect.}
\label{fig:agentroles}
\end{figure}

\paragraph{Specialist-role comparison.}\label{sec:sentiment}
The analyst-only arms enable one role at a time against the all-fallback baseline.
They are configuration comparisons, not additive decompositions of portfolio
influence. Figure~\ref{fig:agentroles} shows all daily allocation distances and
the logical request load.

\begin{figure}[htbp]
\centering
\includegraphics[width=\linewidth]{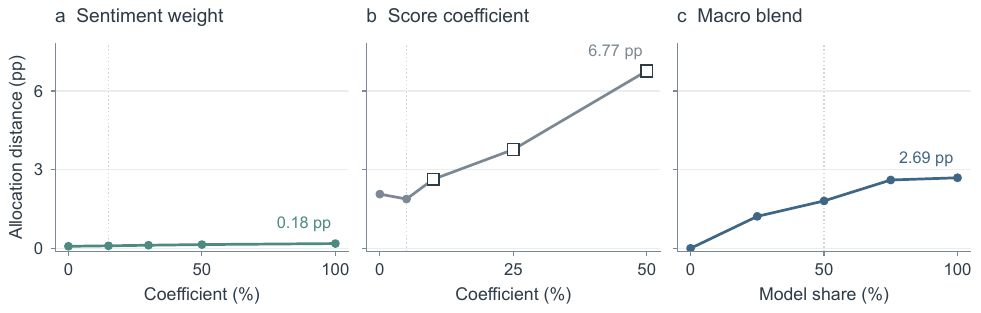}
\caption{Parameter sweeps on retained outputs, with a common vertical scale. Open squares mark membership changes; vertical dotted lines mark the operational allocation coefficients. The x-axes vary the sentiment coefficient, composite-score coefficient before normalization, and macro-model blend share. Permitted actions remain fixed. Source: \protect\path{data/authority_sweep/authority_sweep.csv}.}
\label{fig:authority}
\end{figure}

\paragraph{Allocation coefficient sweeps.}
We re-run the simulator along three parameter axes (Figure~\ref{fig:authority}). Sentiment's analyzer weight takes values 0, 0.15, 0.3, 0.5, and 1, with the macro channel at \Lz{}. The composite-score coefficient takes values 0, 0.05, 0.1, 0.25, and 0.5 under full \Lo{}. The model's share of the macro blend takes values 0, 0.25, 0.5, 0.75, and 1, with sentiment at \Lz{}. That sweep is exact at 0 and 0.5; macro-dependent trend scores are linearly projected elsewhere. Distance at sentiment weight 0 is nonzero because \Lz{} itself carries the lexicon at 0.15.

\begin{figure}[htbp]
\centering
\includegraphics[width=\linewidth]{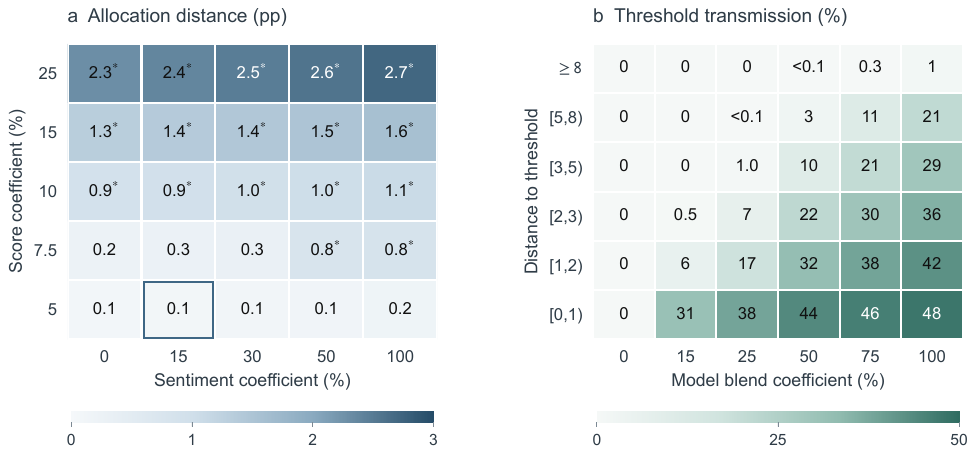}
\caption{(a) Allocation distance from \Lz{} as sentiment and composite-score coefficients vary. The outlined cell marks the operational parameters; an asterisk marks membership changes. (b) Transmission $T$ in the synthetic threshold consumer. Each panel has its own sequential color scale and units. Positive rates below 0.1\% are marked explicitly. Source tables: \protect\path{data/authority_sweep/sentiment_portfolio_interaction.csv} and \protect\path{data/synthetic/phase.csv}.}
\label{fig:phase}
\end{figure}

A two-way sweep crosses sentiment weight with score coefficients of 0.05--0.25 (Figure~\ref{fig:phase}a): at 5\% no setting flips membership; at 7.5\% flips appear only once sentiment weight reaches 0.5; at 10\% or more the same nine dates flip even with sentiment disabled.

The nine membership changes follow the closed-loop path through holdings. A targeted replay check (\protect\path{harness/closed_loop_reach.py}) places all nine dates in 2026-08-11 to 08-21, inside the 15-date window in which the rank buffer passes the held book back into selection, and none before the buffer existed. Same-date selection reads momentum rank and the held set only, so $R$ from the composite to same-date membership is 0 at every setting; what the sweep raises is the gain on the one-date-later path. Paired-bootstrap return intervals (10{,}000 draws, seed 0) include zero at every setting (\protect\path{data/authority_sweep/}). For \Lo{} against \Lz{}, circular block resampling with block lengths 5 and 10 gives $[-3.9,+3.0]$ and $[-3.4,+2.0]$\pp{}.

\paragraph{Synthetic cash-threshold test.}
To check how threshold sensitivity behaves without trading inputs, \protect\path{harness/causal_reach_synthetic.py} draws $n=80{,}000$ fallback scores $b\sim U(0,100)$ and model outputs $z=\mathrm{clip}(b+\Delta,0,100)$ with $E[|\Delta|]=5$ before clipping. It applies $u=\alpha z+(1-\alpha)b$ and the four-level cash quantizer. Transmission $T$ at $\alpha=0.15$ is 2.3\% overall, with no observed changes at $d\ge 3$. At $\alpha=0.5$, it falls from 44\% within one point of a threshold to 0.023\% at $d\ge 8$ (10 of 42,920 draws). These finite-sample rates show concentration near thresholds, not a universal distance beyond which change is impossible.

Selective fallback with an approximate richness schedule (toy failure 0\% at $r\le 5$, rounding the observed 0.04\%, and 27.7\% at $r\ge 16$) yields $\mathrm{corr}(\mathrm{fail},r)=0.31$ and cuts $T$ on rich inputs from 7.8\% to 5.6\%.

Increasing the blend weight in this mean-unbiased synthetic setting leaves mean action almost unchanged (11.26 vs.\ 11.24). It does not reproduce the sizing arm's shift in average cash. This toy statistic is not a return comparison: equal mean exposure can still permit different outcomes through timing and covariance with returns. The experiment is a mechanism check, not a second application.

\begin{figure}[htbp]
\centering
\includegraphics[width=\linewidth]{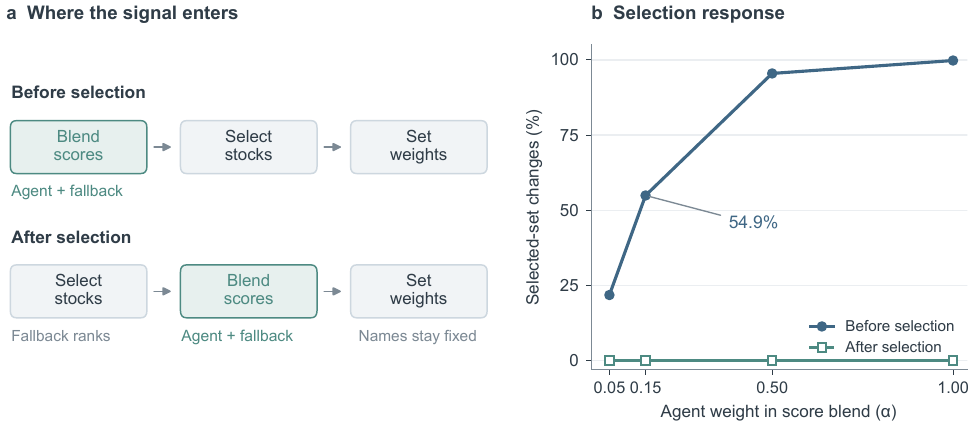}
\caption{Placement changes the available decision. (a) The same agent--fallback score blend enters before or after top-$k$ selection. (b) Selected-set changes across all four retained blend weights, using the same 4,000 synthetic draws, 150 candidates, and ten selections (seed 0). Before-selection scores can change membership; after-selection scores change weights within a fixed set. The toy rule uses normalized squared scores throughout. It isolates a dependency pattern and does not measure trading value.}
\label{fig:placement}
\end{figure}

\paragraph{Model placement in synthetic consumers.}
Three consumers receive the same paired outputs in \protect\path{harness/consumer_topology.py} (Table~\ref{tab:topology}). Consumer A applies the four-level threshold rule to the blend $u=\alpha z+(1-\alpha)b$. Consumer B selects ten of 150 toy candidates using $u$, placing the model \emph{before} selection. Consumer C selects using the fallback $b$ and uses $u$ only for squared-score weights after selection. C therefore has the sentiment component's dependency pattern, while using a different weighting rule.

We draw 4{,}000 universes with $b\sim U(0,100)$ and $E|z-b|=5$, using seed 0. The paired score differences are identical across consumers. The selected set is structurally reachable in B and unreachable in C at each tested nonzero $\alpha$.

In B, the set changes on 22\% of draws at $\alpha=0.05$ and 55\% at $\alpha=0.15$, with means of 0.22 and 0.62 swaps. Sensitivity follows the gap between the 10th- and 11th-ranked fallback scores. At $\alpha=0.15$, the set changes on 65\% of 3{,}093 draws with a gap below one point and 27\% of 709 between one and two. It changes on 4\% of 157 draws between two and three, and on 1 of 41 beyond three.

In C, membership never changes in the tested draws; mean allocation distance is 0.19\pp{} at $\alpha=0.05$ and 0.57\pp{} at $\alpha=0.15$. The toy consumer normalizes squared blended scores for all stock weights and does not reproduce the trading allocation's momentum/composite mixture. A top-$k$ consumer therefore has a decision margin of its own kind, and placement alone decides whether a channel can reach membership.

\paragraph{Fallback sensitivity.}
Realized allocation change is measured against a fallback; structural reachability is a property of the allocation code and admissible outputs. The program \protect\path{harness/fallback_sensitivity.py} fixes the model arm at \Lo{} sentiment-only: model sentiment with quant macro. It compares three baselines that differ only in the sentiment channel (Table~\ref{tab:fallback}). These use the deployed lexicon (\Lz{}), a neutral score without a headline signal (\texttt{L0N}, replayed without model calls), or FinBERT at the same interface. Sign disagreement ranges from 29\% against FinBERT to 99\% against the neutral baseline. The mean composite shift ranges from 3.3 to 4.5 points. Allocation effects remain small: mean half-$\ell_1$ distance is 0.07--0.10\pp{}, and the largest single-name weight change is 0.08--0.09\pp{}. No baseline comparison changes membership. Realized transmission changes with the fallback. Structural reachability instead depends on the admissible output range and consumer code, which remain fixed in this comparison.

\begin{table}[htbp]
\centering
\small
\setlength{\tabcolsep}{2.5pt}
\caption{Fallback sensitivity of the sentiment channel's realized transmission (\protect\path{data/tables/fallback_sensitivity.csv}; 15{,}298 ticker-days, 119 run dates). Output stage: share of ticker-days whose sign relative to neutral differs, mean absolute sentiment and composite differences (0--100 scale). Decision stage: mean half-$\ell_1$ target-book distance and maximum single-name weight change (pp of equity), and membership-flip days. Structural reachability to same-date membership is 0 for every row.}
\label{tab:fallback}
\begin{tabular}{@{}lrrrrrr@{}}
\toprule
Fallback $M_0$ & Sign & $|\Delta$sent$|$ & $|\Delta$comp$|$ & Half-$\ell_1$ & Max $\Delta w$ & Flips \\
\midrule
Lexicon (deployed) & 52\% & 16.0 & 3.8 & 0.098 & 0.086 & 0/119 \\
Neutral (no signal) & 99\% & 17.3 & 4.5 & 0.092 & 0.087 & 0/119 \\
FinBERT & 29\% & 14.4 & 3.3 & 0.070 & 0.082 & 0/119 \\
\bottomrule
\end{tabular}
\end{table}

\begin{table}[htbp]
\centering
\small
\setlength{\tabcolsep}{4pt}
\caption{Deterministic cash sensitivity. Each arm uses cached reconstructed scores and the same portfolio/execution simulator; intervals resample paired daily returns 10{,}000 times (seed 0; \protect\path{harness/cash_sensitivity.py}). The point estimates are monotone in exposure; the intervals are wide.}
\label{tab:cash_sensitivity}
\begin{tabular}{rrrr}
\toprule
Cash & Invested & $\Delta$ ret.\ vs.\ \Lz{} (pp) & 95\% CI \\
\midrule
0\%  & 100\% & $+6.0$ & $[-3.8,+37.7]$ \\
5\%  & 95\%  & $+3.3$ & $[-2.3,+18.2]$ \\
10\% & 90\%  & $+0.5$ & $[-8.1,+6.1]$ \\
15\% & 85\%  & $-3.6$ & $[-25.9,+3.8]$ \\
20\% & 80\%  & $-7.0$ & $[-44.0,+4.7]$ \\
\bottomrule
\end{tabular}
\end{table}

\begin{table}[htbp]
\centering
\small
\setlength{\tabcolsep}{5pt}
\caption{Number of material-change days under alternative half-$\ell_1$ target-book thresholds (\protect\path{harness/materiality_sensitivity.py}). A zero threshold counts every tiny post-selection weight perturbation; the reported conclusions are unchanged across 0.5, 1, and 2\pp{}.}
\label{tab:materiality_sensitivity}
\begin{tabular}{lrrrr}
\toprule
Materiality threshold & 0 & 0.5\pp{} & 1\pp{} & 2\pp{} \\
\midrule
\Lo{} & 119 & 31 & 31 & 31 \\
\Lt{} & 119 & 79 & 79 & 79 \\
Constant 5\% cash & 85 & 85 & 85 & 85 \\
\bottomrule
\end{tabular}
\end{table}

\paragraph{Replay-error perturbation.}
Each \Lo{} sentiment score is perturbed by Gaussian noise whose mean absolute size is $k$ times the observed 9.0-point sentiment MAE ($k=1,2,3$; two seeds each), clipped to $[0,100]$, rebuilt through the production scorer and pushed through the full portfolio and execution simulator. Table~\ref{tab:perturb} reports the transmission against \Lz{} and against the unperturbed \Lo{}.

\begin{table}[htbp]
\centering
\small
\setlength{\tabcolsep}{3pt}
\caption{Perturbed-sentiment arms. Flip days: run dates with any top-ten membership change against \Lz{}; cash days: dates on which the cash level differs from \Lz{}; $\overline{\Delta}$: mean half-$\ell_1$ book distance (pp). Return differences list both seeds; every paired-bootstrap interval endpoint is within 0.5\pp{} of \Lo{}'s.}
\label{tab:perturb}
\begin{tabular}{lrrrrl}
\toprule
Noise ($\times$MAE) & Flips & Cash & $\overline{\Delta}$ vs \Lz{} & $\overline{\Delta}$ vs \Lo{} & Return vs \Lz{} (pp) \\
\midrule
0 (\Lo{}) & 0 & 31 & 1.88 & 0 & $-0.9$ \\
1 & 0 & 31 & 1.89 & 0.04 & $-1.0$, $-0.9$ \\
2 & 0 & 31 & 1.90 & 0.08 & $-1.1$, $-0.9$ \\
3 & 0 & 31 & 1.91 & 0.11 & $-1.2$, $-1.1$ \\
\bottomrule
\end{tabular}
\end{table}

\paragraph{State-conditioned one-step comparison.}
The program \protect\path{harness/conditioned_compare.py} recomputes the \Lz{} and \Lo{} targets on every run date from the same starting state. Before the August 11 rule change, selection is independent of prior holdings. On the 15 dates after the change, recorded pre-trade positions of at least 1\% of equity define rank-buffer incumbents. Each target book is marked from that day's open to the next run date's open, so no simulated path accumulates. Over 119 dates: 0 membership-flip days, 31 cash-change days, mean half-$\ell_1$ distance 1.88\pp{}, summed one-step return difference $-0.81$\pp{} with a paired-bootstrap interval of $[-3.23,+1.56]$ (10{,}000 draws, seed 0); on the 15 state-conditioned dates, 0 flips and 3 cash changes.

\begin{figure}[htbp]
\centering
\includegraphics[width=0.76\linewidth]{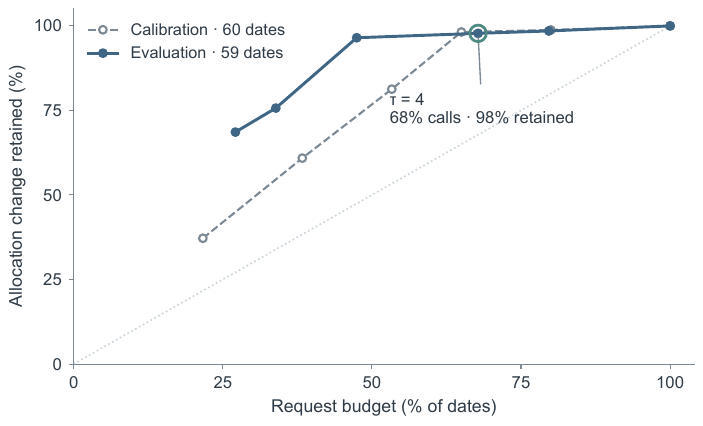}
\caption{Direct allocation-change capture versus selected dates in the 60-date calibration and 59-date evaluation periods. All tested cutoffs are shown. The outlined evaluation point marks $\tau=4$, selected on the calibration period. The diagonal is a proportional-capture reference. Source: \protect\path{data/tables/routing_split.csv}.}
\label{fig:routingfrontier}
\end{figure}
\begin{table}[htbp]
\centering
\small
\setlength{\tabcolsep}{2pt}
\caption{Retrospective routing comparisons. The two 55\% gates select 66 of 119 dates; cash dates count the all-dates arm's 31 changes captured by a gate. Retention is the direct sum of that arm's allocation distances on selected dates divided by its total, not a return or utility measure. The random row gives the 95\% band across 4,000 equal-budget gates; it is not a single simulated policy. The $d_t<4$ row is the full simulated gated arm (98\%); direct capture is also 98\% to rounding.}
\label{tab:routingmain}
\begin{tabular}{@{}llrrr@{}}
\toprule
Policy & Signal & Calls & \shortstack{Cash\\dates} & \shortstack{Allocation change\\retained} \\
\midrule
All macro dates & --- & 100\% & 31/31 & 100\% \\
Input-based gate & input & 55\% & 17/31 & 56\% \\
Random, equal budget & none & 55\% & --- & 40--71\% \\
Margin, closest 55\% & consumer & 55\% & 30/31 & 95\% \\
Margin $d_t<4$ & consumer & 66\% & 31/31 & 98\% \\
\bottomrule
\end{tabular}
\end{table}

\paragraph{Margin routing and temporal stability.}
\protect\path{harness/routing_arm.py} re-simulates \Lo{} (and a macro-only variant) on cached outputs, calling the model only on dates with $d_t<\tau$, $d_t$ the quant macro score's distance to the nearest cash threshold (Table~\ref{tab:routing}). A descriptive full-window heuristic sets $\tau=3$ by rounding up the mean half-gap $|m_t-q_t|/2$ (2.5 points; p90 4.4 and maximum 7.7 over 117 parsed dates). This is not prospective calibration. The split rule picks the smallest $\tau$ that retains 90\% of \Lo{}$-$\Lz{} book distance over the first 60 run dates and evaluates it on the last 59: it selects $\tau=4$, which retains \SPLITTRAIN{} in-sample and \SPLITTEST{} out-of-sample (direct per-day sums; \protect\path{data/tables/routing_split.csv}). Figure~\ref{fig:routingfrontier} plots direct capture separately for calibration and evaluation; threshold selection uses only the stated calibration prefix. Retention in the table is the simulated arm's total book distance from \Lz{} divided by \Lo{}'s; Table~\ref{tab:routingstability} repeats the selection over calibration prefixes of 40--70 dates.

\begin{table}[htbp]
\centering
\small
\setlength{\tabcolsep}{2.5pt}
\caption{Margin router on the reconstruction window: dates with $d_t<\tau$ receive the model call(s), all others run \Lz{}. Cash days: dates whose cash level differs from \Lz{} (\Lo{}: 31). Retained: share of \Lo{}'s total book distance from \Lz{}. Rows without a simulated arm use direct per-day sums of the \Lo{}$-$\Lz{} distances (marked $\dagger$); return differences are counterfactual estimates with paired-bootstrap intervals.}
\label{tab:routing}
\begin{tabular}{@{}lrrrrrrr@{}}
\toprule
$\tau$ & Dates & Calls/d & \$ & Cash d. & Ret. & $\Delta$ (pp) & 95\% CI \\
\midrule
1$^\dagger$ & 29 & 32 & 12 & 15 & 48\% & --- & --- \\
2$^\dagger$ & 43 & 47 & 18 & 21 & 66\% & --- & --- \\
3 (\Lo{}) & 60 & 66 & 25 & 27 & 86\% & $-0.9$ & $[-4.5,+3.5]$ \\
3 (macro only) & 60 & 0.50 & 0.8 & 27 & 85\% & $-0.6$ & $[-4.2,+3.9]$ \\
4 (\Lo{}) & 79 & 87 & 33 & 31 & 98\% & $-0.7$ & $[-4.5,+3.9]$ \\
4 (macro only) & 79 & 0.66 & 1.1 & 31 & 96\% & $-0.7$ & $[-4.5,+4.0]$ \\
5$^\dagger$ & 95 & 104 & 39 & 31 & 99\% & --- & --- \\
\bottomrule
\end{tabular}
\end{table}

\begin{table}[htbp]
\centering
\small
\setlength{\tabcolsep}{3pt}
\caption{Temporal stability of the router rule. For each calibration prefix, $\tau$ is the smallest integer threshold retaining at least 90\% of calibration transmission; metrics are evaluated on the remaining suffix (\protect\path{harness/routing_temporal_stability.py}). Splits overlap and are a stability check, not independent replications.}
\label{tab:routingstability}
\begin{tabular}{rrrrrr}
\toprule
Calibration & Evaluation & $\tau$ & Calls & Transmission & Cash dates \\
\midrule
40 & 79 & 4 & 62.0\% & 96.3\% & 13/13 \\
50 & 69 & 4 & 63.8\% & 96.7\% & 12/12 \\
60 & 59 & 4 & 67.8\% & 97.8\% & 12/12 \\
70 & 49 & 4 & 77.6\% & 98.6\% & 11/11 \\
\bottomrule
\end{tabular}
\end{table}

\paragraph{Corrected-calendar re-issue.}
The primary macro prompts carried the August upcoming-release list on every date. The sensitivity check replaces it with the calendar from Appendix~D, filtered to releases still upcoming at each run date. All 119 calls were reissued once on September 2, 2026; 118 parsed (\protect\path{harness/calendar_check.py}, \protect\path{data/tables/calendar_check.json}). Cash targets can be compared directly from the rounded score blends without a portfolio replay.

The corrected-calendar responses change cash against \Lz{} on 29 dates, compared with 31 in the primary replay. The two sets of cash targets differ on six dates; on 27 dates both change in the same direction. Score correlation is 0.97, mean absolute difference is 1.1 points, and mean distance from the quant score is 5.0 points in both. The corrected replies give \texttt{scale\_in} on 115 of 118 dates, with stance agreement on 109 of 116 shared parsed dates.

Cash changes remain concentrated near thresholds: 28 of the 29 occur at $d_t<4$, and one at $d_t=4.0$. The change rate is 48\% within one point and 2.5\% at four or more points, compared with 52\% and 0\% in the primary sample. The reissue changes both calendar content and the sampled response, so it does not isolate the calendar's causal effect.

\begin{figure}[htbp]
\centering
\includegraphics[width=\linewidth]{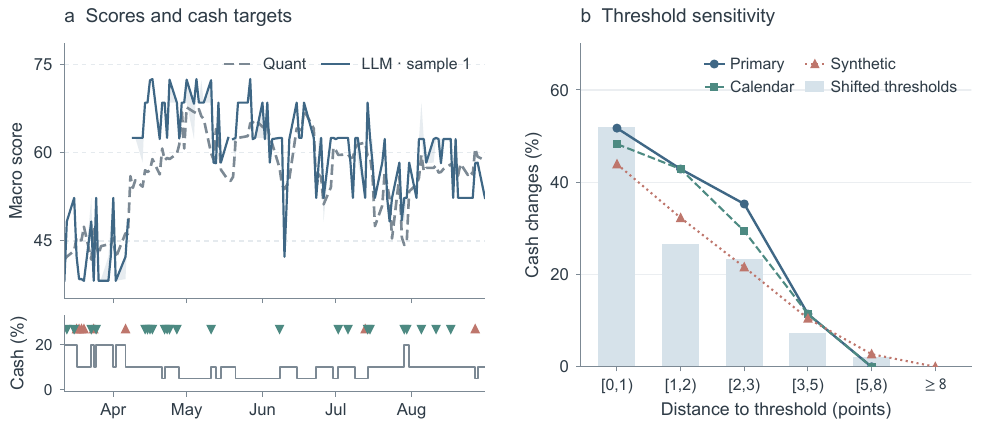}
\caption{Why macro scores change cash. (a) Primary model scores, deterministic scores, and cash targets. Missing replies remain gaps; shading spans two parsed samples, not a confidence interval. Triangles mark more or less advisory cash. (b) Cash-change rates by distance to a threshold, including 819 date--boundary-shift pairs and the primary, corrected-calendar, and synthetic comparisons.}
\label{fig:margin}
\end{figure}

\paragraph{Boundary intervention.}
We shift all three cash thresholds jointly by $\delta=0,\pm2.5,\pm5,\pm7.5$ points. On the 117 parsed dates, we recompute cash for the quant score and the rounded 50/50 blend, then group changes by distance to the nearest shifted threshold (Table~\ref{tab:boundary}). The raw model score differs from the quant score by 5.0 points on average, moving the blend by about 2.5. The resulting rates show how the cash rule converts those score differences into allocation changes.

\begin{table}[htbp]
\centering
\small
\setlength{\tabcolsep}{5pt}
\caption{Share of date--threshold pairs on which the LLM blend changes the cash level, by distance of the quant score to the nearest threshold, pooled over seven threshold shifts.}
\label{tab:boundary}
\begin{tabular}{lrrr}
\toprule
$d_t$ (points) & Date-shift pairs & Cash differs & Share \\
\midrule
$[0,1)$ & 98 & 51 & 52\% \\
$[1,2)$ & 109 & 29 & 27\% \\
$[2,3)$ & 90 & 21 & 23\% \\
$[3,5)$ & 229 & 17 & 7\% \\
$[5,8)$ & 280 & 6 & 2\% \\
$\ge 8$ & 13 & 0 & 0\% \\
\bottomrule
\end{tabular}
\end{table}

\clearpage
\section{Absolute Performance of the Replay Arms}
\label{app:arms}
The original arm summaries below use the first recorded close as the return
origin. They retain 118 close-to-close intervals and omit initial-day P\&L.
The complete initial-capital comparisons in the main text and
Appendix~\ref{app:performance} use a different return origin; their figures
should not be read as numerically interchangeable.
Table~\ref{tab:arms} lists the absolute return, Sharpe ratio, and maximum drawdown of every arm in Table~\ref{tab:where}, plus the FinBERT substitution and the two gated arms of Appendix~D. The arms are reported for completeness; Table~\ref{tab:outcomes} gives the paired differences used in the cash-policy comparison. \Lth{} is not a separate row because its clamp never fired (0/117), so its decisions coincide with \Lt{}'s. The paired intervals are wide relative to the observed differences, so this window does not establish policy equivalence or a profitable model signal.

\begin{table}[htbp]
\centering
\small
\setlength{\tabcolsep}{2pt}
\caption{Absolute performance of the replay arms over 119 recorded run dates 2026-03-12 to 2026-08-31 (last-completed daily bars 03-11 to 08-28; 5~bp per side; \protect\path{data/replay/arms_metrics.csv}). Return is compounded from the 118 close-to-close returns, excluding the initial funding-to-first-close move. Sharpe uses rf $=0$ and annualized daily returns. The principal paired-return intervals include zero (Table~\ref{tab:outcomes}).}
\label{tab:arms}
\begin{tabular}{@{}lrrr@{}}
\toprule
Arm (Table~\ref{tab:where}) & Return (\%) & Sharpe & Max DD (\%) \\
\midrule
\Lz{} Quant-only & 59.9 & 2.09 & $-26.4$ \\
\Lo{} Advisory & 59.0 & 2.06 & $-26.7$ \\
\quad \Lo{} sentiment-only & 59.7 & 2.09 & $-26.4$ \\
\quad \Lo{} macro-only & 59.2 & 2.06 & $-26.7$ \\
\quad FinBERT at the \Lo{} interface & 59.6 & 2.08 & $-26.3$ \\
\Lt{} Sizing & 63.0 & 2.11 & $-26.8$ \\
Constant 5\% cash & 63.2 & 2.12 & $-26.7$ \\
Gate $g_t$ on \Lo{} & 59.5 & 2.07 & $-26.5$ \\
Random gate, equal budget & 60.0 & 2.09 & $-26.5$ \\
\bottomrule
\end{tabular}
\end{table}

\clearpage
\section{Agreement Between Replay and Account Records}
\label{app:faith}
\begin{figure}[htbp]
\centering
\includegraphics[width=0.52\linewidth]{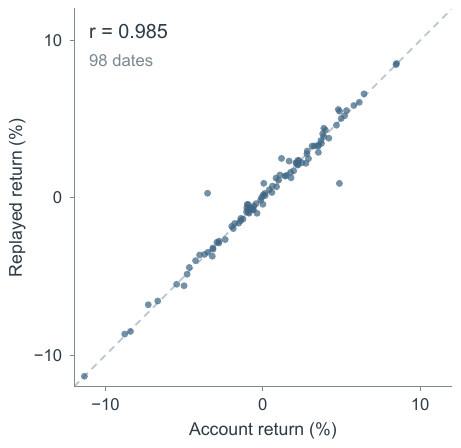}
\caption{Daily returns of the \Lz{} replay against the account over the 98 dates after the no-fill gap. The dashed line denotes equal returns, with both axes in percent.}
\label{fig:equity}
\end{figure}
We assess reconstruction fidelity against the deterministic account and logged runs, then examine transmission under a sentiment-score perturbation. The account record contains 1,073 orders and 1,064 fills over the 119 run dates. No fills occurred during seventeen sessions from March 18 to April 10, when only a report-only local job ran.

Against 32 logged production runs the reconstructed composite score has a mean absolute error of 3.3 points on a 0--100 scale (Spearman 0.948 over 4{,}934 ticker-days), and on 21 dates with real pre-trade holdings available for conditioning, the reconstructed top ten overlaps the logged top ten by 9.5 names.

Daily account and replay returns have Pearson correlation 0.985 over the 98 dates after the gap (Figure~\ref{fig:equity}). The reconstruction reproduces 77\% of account fills when matched by symbol, side, and day over the 83-date daily-trading window, and 75\% over the full pilot. A simulation seeded from the recorded book on August 11 has return correlation 0.988 and a 0.6\pp{} return gap. Full-path replay return exceeds account return by 25.8\pp{} over their overlapping dates; the no-fill gap and residual holdings retained until August contribute to the difference.

Sentiment has the largest component MAE, 9.0 points. Production used ten Yahoo headlines each morning; the reconstruction substitutes Benzinga headlines. The additive-noise check in Appendix~F tests allocation sensitivity at up to three times this measured error, rather than all reconstruction errors.

The validation record (\protect\path{data/l0/validation.md}) documents two defects corrected during reconstruction: a VIX cache exposed future values to historical dates, and pending orders could allow duplicate fills. The executor now aborts runs with a pending order.

\clearpage
\section{Additional Comparison Details}
\label{app:details}
\begin{table}[H]
\centering
\small
\caption{Experimental comparisons and their purposes. Full-path replays evolve their own holdings; an additional same-state comparison holds starting holdings fixed. Synthetic placement tests use a separate toy allocation rule. Cash controls approximate exposure rather than matching the model's full action set.}
\label{tab:procedure}
\begin{tabular}{@{}L{0.18\linewidth}L{0.38\linewidth}L{0.36\linewidth}@{}}
\toprule
Comparison & What is varied & Main readouts \\
\midrule
Model substitution & Model versus deterministic fallback with the same market record and rules & Selected stocks, cash, target-allocation distance \\[3pt]
Rule interventions & Score coefficients, cash boundaries, invocation gates; synthetic placement & Dependence on rule settings, changes retained per request budget \\[3pt]
Cash policies & Model-directed sizing versus constant-cash controls & Exposure, paired returns, uncertainty \\[3pt]
Response formats & Contract and token cap on retained historical inputs & Parsing, usable fields, score and action stability \\
\bottomrule
\end{tabular}
\end{table}

\subsection{Configurations and Protocol}
The primary comparisons use the first of two responses per historical prompt.
Run dates span March 12--August 31, 2026; the last completed daily bars span
March 11--August 28, and run-date opening data are also used. Eligibility and
documented rule changes are reconstructed from code and logs. The complete
historical strategy remains unavailable. Table~\ref{tab:where} gives intervention
permissions, allocation summaries, and logical prompt costs. Costs use rounded
list-price estimates for one sample and omit uncounted retries.

\begin{table}[htbp]
\centering
\small
\setlength{\tabcolsep}{4pt}
\caption{Intervention arms on 119 reconstructed run dates. Allocation distance includes cash and is measured against \Lz{} in percentage points (pp) of equity; the number of dates exceeding 1\pp{} is unchanged at thresholds of 0.5 and 2\pp{}. Prompts/date are average logical prompt tasks, rounded; cost is a rounded list-price estimate for one sample over the window (USD), excluding uncounted retries. These are replay comparisons, not model-directed historical account trades.}
\label{tab:where}
\begin{tabular}{@{}L{3.2cm}L{4.2cm}rrrr@{}}
\toprule
 & \multicolumn{1}{c}{Agent role} & \multicolumn{2}{c}{Allocation effect} & \multicolumn{2}{c}{Cost} \\
\cmidrule(lr){2-2}\cmidrule(lr){3-4}\cmidrule(lr){5-6}
Arm & permitted intervention & \shortstack{Days $>1$\pp{}\\allocation change} & \shortstack{Mean allocation\\distance (pp)} & Prompts/date & \$ \\
\midrule
\Lz{} Quant-only \emph{(operational rule)} & nothing & 0 & 0 & 0 & 0 \\
\Lo{} Advisory \emph{(as designed)} & sentiment coefficient 0.15; half of macro & 31 & 1.9 & 131 & 50 \\
\quad \Lo{} sentiment-only & sentiment only & 0 & 0.1 & 130 & 48 \\
\quad \Lo{} macro-only & macro only & 31 & 1.8 & 1 & 2 \\
\Lt{} Sizing & \Lo{} + stance $\to$ cash 0--40\% & 79 & 4.4 & 131 & 50 \\
Constant 5\% cash & exposure control; no LLM & 85 & 4.6 & 0 & 0 \\
\bottomrule
\end{tabular}
\end{table}

\subsection{Portfolio Distance and Component Comparisons}
For date $t$, $w^1_t$ and $w^0_t$ denote two target portfolios, including cash,
in percentage points of equity. Their allocation distance is
\begin{equation}
D_t=\tfrac12\sum_i |w^1_{ti}-w^0_{ti}|.
\label{eq:distance}
\end{equation}
When targets sum to 100\%, this is the percentage of equity that must move
between them. We also record selected-stock and cash changes. A 1\pp{} cutoff
separates larger changes from small weighting adjustments. Full-path arms evolve
their own holdings; the one-step comparison instead fixes starting holdings.
Individual-role distances need not sum to the two-role distance.

\subsection{Workflow Notation}
At date $t$, $x_t$ denotes available observations and $h_t$ starting holdings.
Task $j$ receives inputs $x_{jt}$. Its model output is $A_j(x_{jt})$, its fallback
$B_j(x_{jt})$, and its pre-request call gate $g_{jt}$. The target portfolio,
including cash, is produced by deterministic coordinator $C_{\theta}$:
\begin{equation}
\begin{aligned}
\widetilde z_{jt}
&=\begin{cases}
A_j(x_{jt}), & g_{jt}=1\text{ and the reply is usable},\\
B_j(x_{jt}), & \text{otherwise},
\end{cases}\\[3pt]
w_t&=C_{\theta}\!\left(x_t,h_t,\{\widetilde z_{jt}\}_j\right).
\end{aligned}
\label{eq:workflow}
\end{equation}
The configuration $\theta$ specifies placement and allocation authority. Execution
turns targets into orders, with resulting holdings feeding later decisions.
This notation describes the compared configurations rather than a learned policy.

\paragraph{Exact cash-rule example.}
On March 13, the deterministic macro score was 42.3 and the model score 48.3.
The equal blend, 45.3, crossed the boundary at 45 and reduced target cash from
20\% to 10\%. Comparisons retain production rounding. If $Q$ maps a score to
cash and $r$ rounds to one decimal, the cash condition is
$Q(r((q_t+m_t)/2))\ne Q(r(q_t))$.

\subsection{Descriptive Signal Checks}
FinBERT is a financial sentiment classifier \citep{araci2019finbert}. The retained
comparison evaluates its sign labels alongside model and lexicon labels; it does
not rank complete trading policies.
\begin{table}[htbp]
\centering
\small
\caption{Descriptive sentiment checks against forward-return direction, in percent. Intervals use a circular block bootstrap over run dates (block length five, 2,000 draws). Sample counts exclude unavailable or neutral signals; horizons overlap. This tests neither a selected trading strategy nor equal performance on a shared evaluation sample. It is separate from the paired allocation comparison.}
\label{tab:signalvalue}
\begin{tabular}{@{}lrrrr@{}}
\toprule
 & \multicolumn{2}{c}{One run date ahead} & \multicolumn{2}{c}{Five run dates ahead} \\
\cmidrule(lr){2-3}\cmidrule(lr){4-5}
Signal & Ticker-days & Accuracy [95\% CI] & Ticker-days & Accuracy [95\% CI] \\
\midrule
LLM sentiment & 14,371 & 49.4 [47.4, 51.4] & 12,764 & 50.1 [47.1, 53.2] \\
Lexicon & 10,672 & 49.4 [48.1, 50.8] & 9,679 & 49.3 [47.0, 52.0] \\
FinBERT & 14,460 & 49.2 [48.1, 50.2] & 12,848 & 49.5 [47.8, 51.2] \\
\bottomrule
\end{tabular}
\end{table}
These sign checks are separate from the allocation comparisons. The signals
have different usable observations, exclude neutral or unavailable labels, and
do not include a selected trading strategy or trading costs. An always-positive
label scores 51.4--51.8\% on the one-date samples and 53.8\% on the five-date
samples. The estimates provide no positive evidence for these tested signals as
incremental trading inputs.

\subsection{Return Estimates}
The retained table below uses the original 118-interval close-to-close convention.
Table~\ref{tab:fullpolicy} reports the updated comparison including the initial
trading interval. Both conventions retain wide intervals containing zero.
\begin{table}[htbp]
\centering
\small
\caption{Simulated outcome differences against \Lz{}. Intervals resample 118 paired close-to-close returns from 119 run dates (10,000 draws, seed 0). The constant-cash rule approximately matches \Lt{}'s exposure, but has a narrower action set. Similar point estimates and intervals containing zero do not establish policy equivalence or absence of model information.}
\label{tab:outcomes}
\begin{tabular}{@{}lrr@{}}
\toprule
Arm vs.\ \Lz{} & $\Delta$ return (pp) & 95\% CI \\
\midrule
\Lo{} Advisory & $-0.9$ & $[-4.7,+3.6]$ \\
\Lt{} Sizing & $+3.1$ & $[-2.4,+17.4]$ \\
Constant 5\% cash & $+3.3$ & $[-2.3,+18.2]$ \\
\bottomrule
\end{tabular}
\end{table}

\subsection{Synthetic Placement}
\begin{table}[htbp]
\centering
\small
\setlength{\tabcolsep}{2pt}
\caption{Synthetic mechanism check with the same paired inputs and blend weight $\alpha=0.15$. Reachability $R$ refers to the threshold action in A and selected-set membership in B/C. Changes is the percentage of draws with an action or selected-set change; swaps is the mean number of names exchanged; distance is mean allocation distance in pp. Allocation weights can change in C even though membership cannot.}
\label{tab:topology}
\begin{tabular}{@{}lcrrrr@{}}
\toprule
Consumer & $R$ & $\alpha$ & Changes & Swaps & Dist. \\
\midrule
A threshold & 1 & 0.15 & 2.1\% & --- & --- \\
B top-$k$, pre-selection & 1 & 0.15 & 54.9\% & 0.62 & 6.4 \\
C top-$k$, post-selection & 0 & 0.15 & 0 & 0 & 0.57 \\
\bottomrule
\end{tabular}
\end{table}
The placement comparison uses the same 4,000 synthetic draws at each blend
coefficient, 150 items per draw, and ten selections. It is distinct from the
80,000-draw scalar-threshold experiment. Before-selection scores can alter
membership; after-selection scores still alter weights within the fixed set.

\clearpage
\section{Portfolio Performance: Definitions and Complete Summaries}
\label{app:performance}
\subsection{Windows, Prices, and Return Origins}
The recorded account window starts at the first positive funded equity point
converted to New York dates. The account value and all its benchmark series
share March 11--August 28. Earlier documentation used the raw UTC date for the
benchmark start, producing a different SPY return. The present comparison uses
the same start and end marks for all four account-window series.

The initial basket holds cash until the first fill date, then buys equal dollar
amounts of the available first-day purchase names at adjusted closes. Its
composition and entry rule differ from the active account, so it is descriptive
context rather than an exposure-matched rebalancing ablation.

Headline reconstructed returns use initial capital before the first trade:
\[
100\left(V_T/V_{\mathrm{initial}}-1\right),\qquad V_{\mathrm{initial}}=\$100{,}000.
\]
This includes the first session's P\&L. The original arm and paired-return tables
retain their first-close origin for traceability. Normalizing each arm at its
first close hides differing initial-day losses, so that convention is not used
for the new headline curves.

Backtest SPY and QQQ buy fractional shares at the first adjusted opening price,
with a five-basis-point entry cost, and use adjusted closes thereafter. Benchmark
and strategy positions remain marked at the endpoint without liquidation costs.
The account benchmarks use the retained adjusted-close series; no hypothetical
five-basis-point charge is added to observed broker marks. Annualized volatility
and zero-rate Sharpe use sample daily-return standard deviation and 252 sessions.
These statistics summarize the observed paths, not their sampling uncertainty.
The replay adds no financing charge beyond its notional trading-cost assumption;
earlier execution versions can permit margin.

\subsection{Execution-Time Sensitivity}
The retained score reconstruction uses later intraday stubs in the earlier
local-run window, while its simulator always fills at the session open. The
original same-session curves therefore admit a timing inconsistency. The
additional program \protect\path{harness/lagged_execution.py} maps every score
and macro record to the next observed trading date, recomputes target portfolios
against each arm's evolving holdings, and applies the existing versioned
execution rules. Scores from the last date are not traded; the first session
holds initial cash. The map is strictly forward in time.

This comparison changes assessment age and can change holdings feedback. It does
not isolate a fill-price correction or recreate the account. It uses cached
outputs without new model calls. Restricted raw prices and the private strategy
are required to regenerate these paths; the normalized curves, per-arm records,
summary, and source hashes are retained. All three primary configurations and
all five constant-cash sensitivities were run, rather than selecting a favorable
lagged configuration.

\begin{table}[htbp]
\centering\small
\caption{Complete return and risk summaries. Return and maximum drawdown are in percent; Sharpe uses a zero risk-free rate. The lower groups are simulated policies. SPY and QQQ are identical across the two backtest timing settings and are printed once. Different evidence groups have different date windows and should not be combined as one realized trajectory.}
\label{tab:fullperformance}
\begin{tabular}{@{}lrrr@{}}
\toprule
Series & Return & Max drawdown & Sharpe \\
\midrule
\multicolumn{4}{@{}l}{\emph{Recorded account window: March 11--August 28}} \\
Paper account & +34.5 & -26.8 & 1.46 \\
SPY & +14.4 & -6.3 & 2.10 \\
QQQ & +18.2 & -11.2 & 1.66 \\
Initial basket & +18.5 & -11.2 & 1.67 \\
\addlinespace[4pt]
\multicolumn{4}{@{}l}{\emph{Same-session reconstruction: March 12--August 31}} \\
L0 & +56.4 & -26.4 & 1.99 \\
L1 & +55.5 & -26.7 & 1.96 \\
L2 & +58.7 & -26.8 & 2.00 \\
SPY & +14.8 & -5.6 & 2.17 \\
QQQ & +19.1 & -11.2 & 1.74 \\
\addlinespace[4pt]
\multicolumn{4}{@{}l}{\emph{Next-session assessments: March 12--August 31}} \\
L0 & +72.0 & -26.5 & 2.34 \\
L1 & +71.0 & -27.4 & 2.32 \\
L2 & +78.3 & -27.3 & 2.41 \\
\addlinespace[4pt]
\bottomrule
\end{tabular}
\end{table}

\begin{table}[htbp]
\centering\small
\caption{Every configuration run in the next-session sensitivity, with its original same-session reference. Values are complete initial-capital returns (percent), not paired confidence estimates. Constant-cash arms use baseline stock scores. This table preserves the full tested set, including constant policies with higher returns than the agent sizing arm.}
\label{tab:timingpolicies}
\begin{tabular}{@{}lrr@{}}
\toprule
Policy & Same-session reconstruction & Next-session assessments \\
\midrule
L0 & +56.38 & +72.01 \\
L1 & +55.54 & +71.05 \\
L2 & +58.72 & +78.30 \\
Constant 0\% & +61.32 & +81.99 \\
Constant 5\% & +58.95 & +77.14 \\
Constant 10\% & +56.44 & +72.89 \\
Constant 15\% & +52.69 & +68.72 \\
Constant 20\% & +49.53 & +63.75 \\
\bottomrule
\end{tabular}
\end{table}

\subsection{Full-Window Policy Uncertainty}
The updated paired comparison includes 119 return intervals, starting with the
initial open-to-close interval. Each bootstrap draw resamples paired date indices
for the policy and baseline, compounds each selected return path, then subtracts
its return. The circular block sensitivity samples consecutive five-date blocks
until the same number of intervals is reached. Both use 10,000 draws and seed 0.
They describe conditional within-window variation and do not correct input,
model-knowledge, or execution defects. All displayed intervals include zero.

\begin{table}[htbp]
\centering\small
\caption{Policy return differences versus \Lz{}, including the initial trading interval (pp). Both 95\% interval constructions retain pairing. These estimates use the original same-session reconstruction, not the next-session sensitivity.}
\label{tab:fullpolicy}
\begin{tabular}{@{}lrrr@{}}
\toprule
Policy & Difference & Paired-date interval & Five-date block interval \\
\midrule
L1 Advisory & -0.84 & $[-4.63,+3.48]$ & $[-3.81,+3.01]$ \\
L2 Sizing & +2.34 & $[-3.10,+15.58]$ & $[-3.05,+14.76]$ \\
Cash 0\% & +4.94 & $[-4.32,+34.65]$ & $[-3.80,+31.36]$ \\
Cash 5\% & +2.57 & $[-2.93,+16.18]$ & $[-2.91,+15.06]$ \\
Cash 10\% & +0.06 & $[-8.50,+5.41]$ & $[-7.83,+5.99]$ \\
Cash 15\% & -3.69 & $[-25.38,+3.58]$ & $[-22.53,+3.28]$ \\
Cash 20\% & -6.86 & $[-42.20,+4.73]$ & $[-37.50,+4.08]$ \\
\bottomrule
\end{tabular}
\end{table}

\end{document}